\documentclass{article} 
\usepackage{iclr2027_conference,times}
\usepackage{needspace}
\usepackage{graphicx}
\usepackage[table]{xcolor}
\usepackage{float}
\usepackage{graphicx,wrapfig}
\usepackage{booktabs}
\usepackage{multirow}
\usepackage{array}
\usepackage{enumitem}
\usepackage{fvextra}
\usepackage[most]{tcolorbox}
\usepackage[scaled=0.95]{inconsolata}

\definecolor{promptviolet}{HTML}{B052B4}

\iclrfinalcopy

\newtcolorbox{eninputbox}[1][]{
    enhanced,
    colframe=promptviolet,
    colback=promptviolet!6,
    colbacktitle=promptviolet,
    title={Chinese Prompt Translation},
    coltitle=white,
    fonttitle=\bfseries\fontsize{9}{10}\selectfont,
    boxrule=0.7pt,
    arc=2mm,
    outer arc=2mm,
    left=1pt,
    right=1pt,
    top=1pt,
    bottom=1pt,
    toptitle=1.2pt,
    bottomtitle=1.2pt,
    before skip=5pt,
    after skip=5pt,
    #1
}

\usepackage{amsmath,amsfonts,bm}

\def\eqref#1{equation~\ref{#1}}

\def\1{\bm{1}}

\DeclareMathAlphabet{\mathsfit}{\encodingdefault}{\sfdefault}{m}{sl}
\SetMathAlphabet{\mathsfit}{bold}{\encodingdefault}{\sfdefault}{bx}{n}

\usepackage{hyperref}
\usepackage{url}
\usepackage{xcolor}
\definecolor{mydarkblue}{RGB}{0,70,140}

\hypersetup{
    colorlinks=true,
    linkcolor=black,
    citecolor=mydarkblue,
    urlcolor=mydarkblue
}

\title{RESCUE: Repairing Language Model Errors to Sparse Circuits via Reinforcement Learning}

\author{%
\textbf{%
Chuanpu Liu\textsuperscript{1}\thanks{Equal contribution.},\enspace
Miao Yu\textsuperscript{2$\ast$},\enspace
Yikai Cai\textsuperscript{1},\enspace
Yuanhe Zhang\textsuperscript{1},\enspace
Zhenhong Zhou\textsuperscript{3}} \\[0.3em]
\textbf{%
Li Sun\textsuperscript{1}\thanks{Corresponding authors.},\enspace
Zuming Jiang\textsuperscript{2},\enspace
Yufei Guo\textsuperscript{4$\dagger$}} \\[0.8em]
\textnormal{%
\textsuperscript{1}Beijing University of Posts and Telecommunications
\quad
\textsuperscript{2}University of Hong Kong} \\[0.2em]
\textnormal{%
\textsuperscript{3}Nanyang Technological University
\quad
\textsuperscript{4}China Aerospace Science and Industry Corporation}
}

\begin{document}
\raggedbottom

\maketitle
\fancyhead{}
\lhead{Preprint}


\begin{abstract}
Large language models (LLMs) exhibit strong general capabilities that mechanistic interpretability has attributed to sparse computational circuits. 
However, existing circuit studies emphasize preserving functionality or explaining safety, leaving the mechanisms underlying failures across a broader range of tasks largely unexplored.
Extending circuit analysis from abilities to errors, we explore the perspective that such failures may likewise arise from erroneous internal computations and that targeted tuning of the corresponding parameters can correct such errors while largely preserving other capabilities.
Motivated by this insight, we introduce \textbf{RESCUE} (\textbf{R}easoning-\textbf{E}rror \textbf{S}parse-\textbf{C}ircuit \textbf{U}ncovering and \textbf{E}diting), a framework that localizes error-associated circuits and surgically repairs them for performance enhancement.
General tasks typically involve multi-step reasoning and long-form generation, where early deviations can cause prefixes to drift from supervised references, leading SFT-based mask optimization to overlook circuits involved in generation-time errors.
RESCUE therefore refines these masks through reinforcement learning with multiple masked-model rollouts, improving their relevance to observed task failures.
Finally, RESCUE introduces a pruning technique and precisely fine-tunes error circuits to correct task failures, thereby translating error localization into a sparse and targeted model update.
We validate RESCUE on heterogeneous repair sets across two domains: \textbf{(1) mathematical reasoning}, identifying a math error circuit of 1.40\% density whose repair raises accuracy from $6.0\% \rightarrow 75.5\%$; and \textbf{(2) medical QA}, where a similarly compact 1.44\% circuit improves repair-set accuracy from $0\% \rightarrow 81\%$.
Our code is available at:
\url{https://github.com/chuanpupig/RESCUE}.
\end{abstract}
\section{Introduction}


Recent advances~\citep{ouyang2022training,rafailov2023direct} have enabled large language models (LLMs) to achieve strong performance across a broad range of tasks. 
Nevertheless, they continue to exhibit recurring failures even on seemingly simple problems, such as elementary arithmetic, limiting their reliability~\citep{zhang2024arithmetic,nikankin2025arithmetic,yu2025survey}. 
Existing analyses and remedies often emphasize data coverage or training objectives~\citep{lin2022truthfulqa,kandpal2023longtail,ji2023hallucination}, but provide limited insight into the internal computations responsible for these errors. 
Meanwhile, growing evidence suggests that specific model behaviors are mediated by sparse internal components and computational circuits~\citep{dai2022knowledge,todd2024function,meng2022locating}. 
Motivated by these findings, we propose that recurring failures may likewise arise from faulty computations concentrated in sparse error-associated circuits, and that localizing them enables effective repair through circuit-restricted tuning with limited impact on other capabilities.

We categorize existing studies related to error attribution into two main research lines: 
\textbf{(I) Task circuits.} Prior work has uncovered sparse circuits underlying capabilities such as factual recall and reasoning~\citep{yao2024knowledge,hong2025implies,hou2023mechanistic}, but has primarily sought circuits that faithfully reproduce or preserve successful task behavior. Their task-specific attribution methods do not readily transfer to error attribution or targeted repair. 
\textbf{(II) Trustworthy circuits.} Safety-oriented research demonstrates that localized components can support selective interventions on alignment~\citep{yu2026safeseek} and backdoor behaviors~\citep{zhao2025understanding,yu2025backdoor}, yet remain focused on safety-specific mechanisms.
Hallucination-attribution work such as H-Neurons~\citep{gao2025hneurons} links erroneous outputs to individual neurons.
However, whether localized interventions can repair a wider range of failures remains unexplored.
Together, there is a lack of methods that can efficiently locate circuits and use them for targeted repair across broader task domains.


\begin{figure}[t!]
    \centering
    \includegraphics[width=\linewidth]{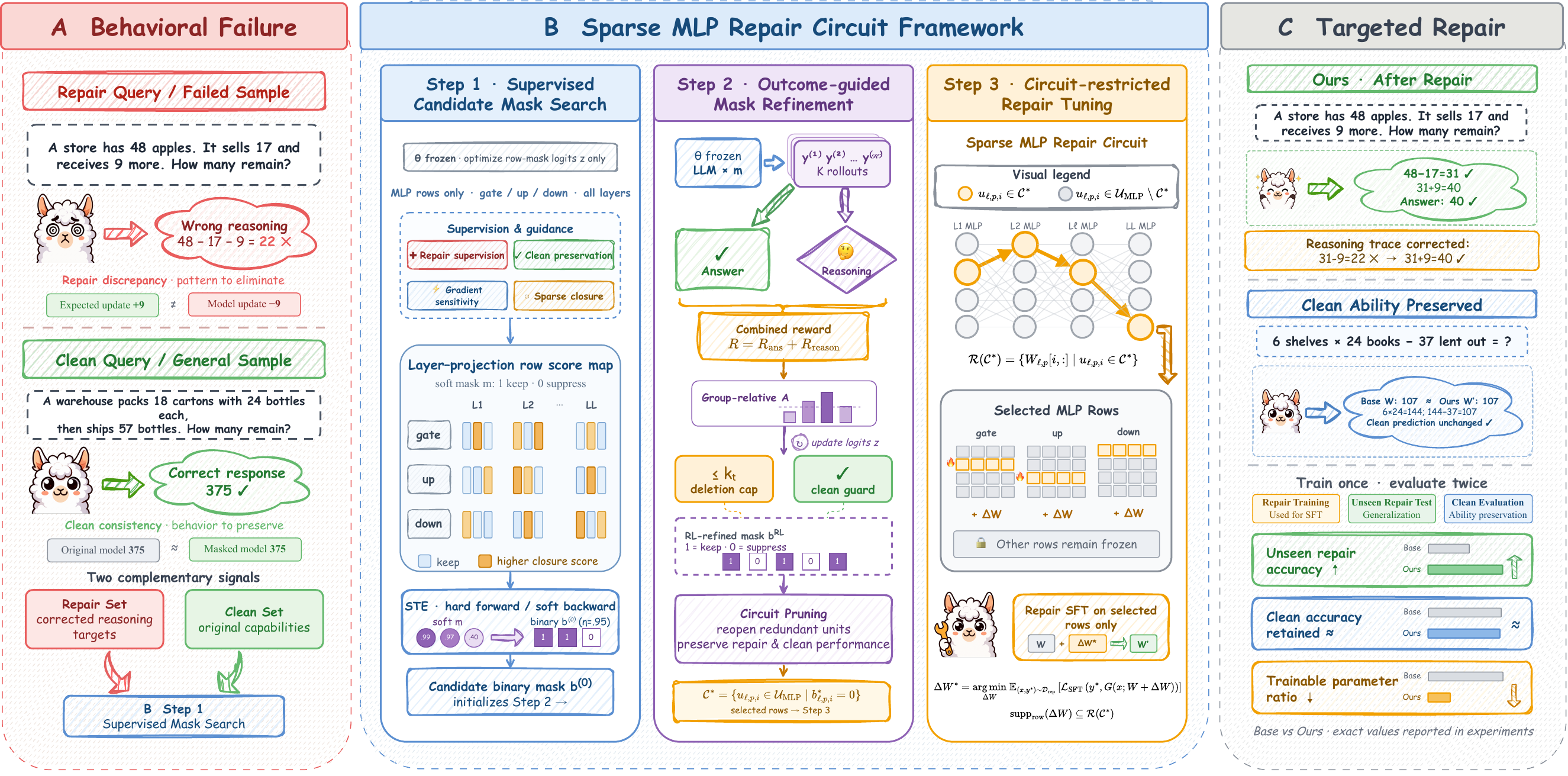}
    \vspace{-2em}
    \caption{Overview of \textbf{RESCUE}, with complementary repair and preservation signals (\textit{\textbf{Left}}), the sparse-circuit localization and tuning framework (\textit{\textbf{Middle}}), and targeted repair generations (\textit{\textbf{Right}}).}
    \label{fig:main}
    \vspace{-13.5pt}
\end{figure}

To bridge these gaps, we propose \textbf{RESCUE} (\textbf{R}easoning-\textbf{E}rror \textbf{S}parse-\textbf{C}ircuit \textbf{U}ncovering and \textbf{E}diting), a framework that localizes sparse circuits associated with task failures and selectively tunes them to improve the corresponding task performance.
RESCUE localizes sparse circuits associated with model failures through a two-stage optimization.
We first isolate error-associated circuits through SFT-based mask optimization using teacher-corrected responses, with model weights frozen.
Yet the mismatch between offline references and free generation may prevent mask optimization from identifying all components contributing to erroneous outputs.
To improve the accuracy and generalizability of error-circuit localization, we refine the candidate masks in the second stage using feedback from multiple rollouts of the masked model.
With a group-relative policy-gradient objective over these multiple generations, RESCUE updates the mask toward components that consistently contribute to failure.
The refined masks are then pruned to obtain the final sparse circuit, and only the corresponding parameters are tuned for targeted repair.

We evaluate RESCUE on Qwen3-8B~\citep{yang2025qwen3} and Llama-3.1-8B-Instruct~\citep{grattafiori2024llama3}.
In the mathematical domain, RESCUE identifies sparse circuits associated with model errors in both models, and then fine-tuning only these localized parameters raises GSM8K~\citep{cobbe2021training} accuracy from $80\% \rightarrow 91\%$ and improves MATH-500~\citep{hendrycks2021math,lightman2023verify} from $75\% \rightarrow 78\%$.
RESCUE outperforms the supervised-only interpretable baseline and vanilla fine-tuning on GSM8K by 4 and 3 percentage points, respectively, highlighting the benefit of refining masks according to response-level task performance.
In the medical domain, RESCUE likewise outperforms vanilla fine-tuning on the MedMCQA~\citep{pal2022medmcqa}
repair sets.
Across all repair settings, RESCUE retains an average of $99.4\%$ of baseline aggregate performance across eight non-target benchmarks, indicating that targeted repair largely preserves general capabilities.
These results demonstrate that attributing task failures to sparse internal mechanisms provides an effective basis for targeted repair while largely preserving non-target capabilities.

In summary, we introduce RESCUE, a generation-guided framework that combines supervised mask optimization, reinforcement learning over multiple masked-model rollouts, and circuit pruning to localize sparse circuits associated with task failures. Their causal effects transfer across input rephrasings, and tuning only the corresponding parameters enables targeted error correction while largely preserving non-target performance, illustrating how circuit attribution can support model repair.
\section{Related Work}

\textbf{Circuit Discovery.}
Circuit discovery explains model behavior through sparse computational subgraphs~\citep{wang2023interpretability}.
Methods span Path Patching and ACDC~\citep{goldowskydill2023path,conmy2023acdc}, scalable Attribution Patching, Edge Pruning, and IBCircuit~\citep{syed2024attribution,bhaskar2024edge,bian2025ibcircuit}, and feature-level Sparse Feature Circuits and Circuit-Tracer~\citep{marks2025sparse,hanna2025circuit}, with applications to performance improvement~\citep{yao2024knowledge,cho2025revisiting,kharlapenko2025scaling}.
Yet heuristic search requires costly repeated interventions, while scalable alternatives target behavior preservation rather than error attribution.
RESCUE instead combines supervised mask optimization with feedback from multiple rollouts, avoiding exhaustive search and enabling generation-aware error localization for targeted repair.

\textbf{Error Attribution.}
Hidden-state methods probe latent knowledge with CCS \citep{burns2023latent}, falsehood and hallucination \citep{azaria2023internal,chen2024inside}, truthfulness \citep{orgad2025know}, and arithmetic errors \citep{sun2025probing}, but decodability does not establish causality.
Component-level work attributes factual errors to attention and MLP mechanisms through SAT Probe \citep{yuksekgonul2024attention}, mechanistic tracing \citep{yu2024mechanistic}, FAITH \citep{yuan2024whispers}, and ReDeEP \citep{sun2025redeep}, while arithmetic studies identify critical heads, MLPs, and circuits \citep{zhang2024arithmetic,hanna2023greater}.
ITI \citep{li2023iti}, TruthX \citep{zhang2024truthx}, and DoLa \citep{chuang2024dola} further use internal signals for inference-time intervention.
\section{Preliminary}

\textbf{LLM Computational Graph.}
Let $M_{\theta}$ denote a language model parameterized by $\theta$.
Following the computational-graph view of language models \citep{yao2024knowledge}, we represent its forward computation as a directed graph $\mathcal{G}=(\mathcal{V},\mathcal{E})$.
At a chosen level of abstraction, each node $v\in\mathcal{V}$ represents an internal computation, while each directed edge $(u,v)\in\mathcal{E}$ specifies how the output of an upstream computation $u$ contributes to a downstream computation $v$.
Given an input sequence $\mathbf{x}$, the complete graph induces the conditional output distribution
$p_{\theta}(\mathbf{y}\mid\mathbf{x};\mathcal{G})$.

\textbf{Circuit Units.}
We use an output-dimension-level granularity for Transformer MLP projections, following prior circuit-attribution work~\citep{geva2022promoting,yu2026safeseek}. Specifically, each output dimension of the gate, up, and down projections is treated as an individual circuit unit. For projection $p\in\{\mathrm{gate},\mathrm{up},\mathrm{down}\}$ in layer $l$, let $\mathbf{W}_{l,p}\in\mathbb{R}^{d_{\mathrm{out}}\times d_{\mathrm{in}}}$ denote its weight matrix. Given an input activation $\mathbf{h}\in\mathbb{R}^{d_{\mathrm{in}}}$, the $j$-th unit produces the corresponding scalar projection output as follows:
\[
n_{l,p,j}(\mathbf{h})
=
[\mathbf{W}_{l,p}\mathbf{h}]_j
=
\mathbf{w}_{l,p,j}^{\top}\mathbf{h},
\]
where $\mathbf{w}_{l,p,j}^{\top}$ is the $j$-th row of $\mathbf{W}_{l,p}$. We index each unit as $u=(l,p,j)$ and denote the set of all such units across MLP projections and Transformer layers by $\mathcal{U}$.

\textbf{Circuit Formulation.}
For a task distribution $\mathcal{D}$, a circuit $\mathcal{C}\subseteq\mathcal{G}$ is a sparse computational subgraph that faithfully retains the task-relevant behavior of the complete model \citep{conmy2023acdc,bhaskar2024edge}.
Circuit discovery can therefore be formulated as
\begin{equation}
\begin{aligned}
\min_{\mathcal{C}\subseteq\mathcal{G}}\quad
&\mathbb{E}_{\mathbf{x}\sim\mathcal{D}}
\left[
D_{\mathrm{KL}}\!\left(
p_{\theta}(\cdot\mid\mathbf{x};\mathcal{G})
\,\Vert\,
p_{\theta}(\cdot\mid\mathbf{x};\mathcal{C})
\right)
\right]
\text{s.t.}\quad
&\frac{|\mathcal{C}|}{|\mathcal{G}|}\leq\gamma ,
\end{aligned}
\label{eq:circuit_formulation}
\end{equation}
where $D_{\mathrm{KL}}$ measures the behavioral discrepancy between the circuit and the complete model, and $\gamma$ specifies the upper bound on the relative circuit size.
Unlike conventional circuits defined as retained subgraphs, RESCUE defines an error-associated circuit as a sparse set of units whose suppression corrects model failures while preserving behavior on clean inputs.
Here, $|\cdot|$ counts computational elements under the granularity adopted by the circuit-discovery method.
\section{RESCUE}
We present \textbf{RESCUE}, a framework for localizing and repairing sparse error-associated circuits in general tasks. We first construct complementary error-correction and clean datasets (Sec.~\ref{sec:data_construction}), then localize circuits through supervised mask search, output-guided reinforcement learning, and pruning (Sec.~\ref{sec:circuit_discovery}), before circuit validation and circuit-restricted tuning (Sec.~\ref{sec:circuit_tuning}).
\subsection{Data Construction}
\label{sec:data_construction}
To construct dual-path supervision for error-circuit discovery, we first run the target model on each question $x$ and partition the resulting responses according to final-answer correctness.
For a correctly answered question $x_j$, we retain the model's original response $y_j$, forming the clean set $\mathcal{D}_{\mathrm{clean}}=\{(x_j,y_j)\}$.
For an incorrectly answered question $x_i$, we retain the model's original erroneous response $y_i^{-}$ and use an external LLM to minimally revise it into a corrected response $y_i^{+}$, forming the error-correction set $\mathcal{D}_{\mathrm{err}}=\{(x_i,y_i^{-},y_i^{+})\}$.
During candidate circuit search, the corrected responses in $\mathcal{D}_{\mathrm{err}}$ provide error-correction supervision, while $\mathcal{D}_{\mathrm{clean}}$ is used to preserve the model's existing correct behavior.
In practice, we use GSM8K~\citep{cobbe2021training} for mathematical reasoning and MedMCQA~\citep{pal2022medmcqa} for medical QA, covering multiple general tasks.

\subsection{Error Circuit Discovery}
\label{sec:circuit_discovery}
The goal of error circuit discovery is to identify a sparse set of units whose suppression corrects erroneous responses while preserving the model's existing correct behavior.
Given the search space $\mathcal{U}$ defined above, RESCUE freezes the original model parameters and assigns each circuit unit $u\in\mathcal{U}$ a trainable mask logit $z_u$.
We use keep-mask semantics, where a mask value of one preserves the unit and a value of zero suppresses its output.
Specifically, we define
\begin{equation}
c_u = \operatorname{clip}_{[0,1]}
\left( 2\sigma\left(\frac{z_0-z_u}{\tau}\right)-1 \right),
\qquad m_u = 1-c_u
\end{equation}
where $c_u$ denotes the closure strength and $\tau$ is the mask temperature.
Initializing $z_u=z_0$ gives $m_u=1$ for all units, leaving the initial model's computation unchanged.
With this initialization, RESCUE first obtains a candidate circuit through reference-based mask optimization, then refines it using generation-level feedback, and finally removes redundant units through circuit pruning.

\textbf{Candidate Circuit Search.}
For each error example $(x_i,y_i^{-},y_i^{+})\in\mathcal{D}_{\mathrm{err}}$, we use the fixed corrected response $y_i^{+}$ as the optimization target, while keeping the model parameters frozen.
We denote the token-level negative log-likelihood on the corrected responses as $\mathcal L_{\mathrm{corr}}$.
To preserve the model's existing correct behavior, we also minimize the token-level negative log-likelihood on the clean responses, denoted by $\mathcal L_{\mathrm{clean}}$.
Let $\mathcal P$ denote the set of MLP layer-projection modules, and let $\mathcal U_q$ denote the units in module $q\in\mathcal P$.
The candidate-search objective is
\begin{equation}
\mathcal L_{\mathrm{cand}} = \mathcal L_{\mathrm{corr}} + \lambda_{\mathrm{clean}}\mathcal L_{\mathrm{clean}} + \lambda_{\mathrm{keep}}
\frac{1}{|\mathcal P|}
\sum_{q\in\mathcal P}
\frac{1}{|\mathcal U_q|}
\sum_{u\in\mathcal U_q}m_u + \mathcal L_{\mathrm{sens}}.
\end{equation}
The third term is a mild keep penalty that encourages mask values to decrease, while the per-module deletion budget introduced below limits the number of closed units.
$\mathcal L_{\mathrm{sens}}$ favors closing units that improve correction while keeping clean-sensitive units open.

Although continuous keep scores enable gradient-based optimization, they do not directly define the discrete intervention required to isolate an error circuit. Binarizing them only after training could introduce a mismatch between mask optimization and circuit extraction. We therefore employ the \textit{Straight-Through Estimator} (STE) throughout candidate search.
At each optimization step, the continuous keep scores are converted into binary keep masks $b_u\in\{0,1\}$ using a threshold $\eta_0$ and a per-module deletion budget $\rho_0$.
Within each module $q$, at most $\lfloor\rho_0|\mathcal{U}_q|\rfloor$ units can be closed.
If more units satisfy $m_u\leq\eta_0$ than allowed by the budget, only those with the smallest keep scores are closed.
The binary masks are used in the forward pass, while gradients are propagated through the continuous keep scores using the straight-through estimator:
\begin{equation}
b_u^{\mathrm{ste}} = m_u+\operatorname{sg}(b_u-m_u),
\qquad
\widetilde u(\mathbf h) = b_u^{\mathrm{ste}}u(\mathbf h).
\end{equation}
Here, $\operatorname{sg}(\cdot)$ denotes the stop-gradient operator. 
In the forward pass, $b_u^{\mathrm{ste}}=b_u$, so the model uses binary masks. 
During backpropagation, gradients flow through $m_u$, allowing the mask logits to be optimized. 
Thus, mask optimization uses the same discrete interventions as circuit extraction.
After optimization, the units with final binary masks $b_u^{(0)}=0$ form the candidate circuit $\mathcal C_{\mathrm{cand}}$.

\textbf{RL-Based Circuit Refinement.}
The reference-based candidate search uses fixed corrected responses and may therefore miss units that sustain errors during free generation.
Initialized from the candidate mask $\mathbf b^{(0)}$, RESCUE addresses this limitation by sampling multiple responses from the masked model and updating only the mask logits using generation-level outcome feedback.

For each error question $x_i$ with ground-truth answer $a_i$, the masked model samples a group of $K$ responses
$\{\hat{y}_{i,k}\}_{k=1}^{K}$.
We score each response based on final-answer correctness and reasoning quality, and compute its group-normalized advantage as
\begin{equation}
r_{i,k} = \lambda_{\mathrm{ans}}r_{\mathrm{ans}}(\hat y_{i,k},a_i) + \lambda_{\mathrm{rea}}r_{\mathrm{rea}}(x_i,\hat y_{i,k},a_i),
\qquad
A_{i,k} = \frac{r_{i,k}-\bar r_i}{\max(s_i,\epsilon)}.
\end{equation}
Here, $r_{\mathrm{ans}}$ measures final-answer correctness, $r_{\mathrm{rea}}$ is provided by an external reasoning evaluator, and $\bar r_i$ and $s_i$ denote the mean and standard deviation of the rewards within the response group. The coefficients $\lambda_{\mathrm{ans}}$ and $\lambda_{\mathrm{rea}}$ balance the two reward components, and $\epsilon$ is a small constant for numerical stability. We then optimize the sequence-level objective
\begin{equation}
\mathcal L_{\mathrm{RL}} = -\mathbb E_i
\left[
\frac{1}{K}
\sum_{k=1}^{K}
A_{i,k}
\frac{1}{T_{i,k}}
\sum_{t=1}^{T_{i,k}}
\log p_{\theta,\mathbf b^{\mathrm{ste}}}
\left( \hat y_{i,k,t} \mid x_i,\hat y_{i,k,<t} \right)
\right].
\end{equation}
This group-relative objective increases the likelihood of higher-reward responses and decreases that of lower-reward responses for the same question.
It therefore refines the masks using multiple free-running outcomes rather than a single fixed response.
During refinement, we retain the STE-based binary masking in Eq. (4) and impose a stage-specific per-module deletion budget.
Thus, the model uses discrete masks during generation while gradients are propagated through the continuous masks.

We further stabilize refinement with an auxiliary negative log-likelihood loss on the corrected response $y_i^{+}$, assigning it a larger weight when the sampled responses receive uniform rewards and the group-relative signal becomes uninformative. 
We also compute a negative log-likelihood loss on $\mathcal{D}_{\mathrm{clean}}$, whose gradient helps preserve the model's existing correct behavior.
When the refinement gradient conflicts with the clean-preservation gradient, we remove its conflicting component before updating the masks.
Mask updates are accepted only when they satisfy both error-correction and clean-performance constraints; otherwise, the model is restored to the last feasible mask.
Once the correction target is reached and the clean-performance constraint is satisfied, we progressively tighten the deletion budget to obtain a smaller binary circuit.

\textbf{Circuit Pruning.}
The circuit returned by RL may contain unnecessarily closed units, which we prune by reopening them.
We rank the closed units by closure strength $c_u$, from weakest to strongest, and test them for reopening in progressively smaller groups.
A group is reopened if the resulting mask preserves error-correction and clean-task performance relative to the RL-refined mask:
\begin{equation}
S_{\mathrm{err}}(\mathbf{b}')
\ge
S_{\mathrm{err}}(\mathbf{b}^{\mathrm{RL}})-\epsilon_{\mathrm{err}},
\qquad
S_{\mathrm{clean}}(\mathbf{b}')
\ge
S_{\mathrm{clean}}(\mathbf{b}^{\mathrm{RL}})-\epsilon_{\mathrm{clean}}.
\end{equation}
Here, $\mathbf b'$ is the proposed mask, and $S_{\mathrm{err}}$ and $S_{\mathrm{clean}}$ denote the accuracies on error and clean examples, respectively.
This process retains only the units whose closure remains necessary.
The remaining closed units form the final error circuit $\mathcal C^*=\{u\in\mathcal U\mid b_u^*=0\}$.

\subsection{Circuit Validation and Tuning}
\textbf{Circuit Ablation Validation.}
\label{sec:circuit_validation}
To assess the effect of suppressing the final error circuit $\mathcal C^*$, we compare the original model $\mathcal G$ with the circuit-ablated model $\mathcal G\setminus\mathcal C^*$. For $\mathcal X\in\{\mathcal G,\mathcal G\setminus\mathcal C^*\}$, we measure normalized final-answer exact-match accuracy on an evaluation set $\mathcal D$ as
\begin{equation}
\operatorname{Score}(\mathcal D;\mathcal X) = \mathbb E_{(x,a)\sim\mathcal D}
\left[ \mathbb I\!\left(
\operatorname{Ans}(\mathcal X(x))=a \right)
\right].
\end{equation}
A higher score for $\mathcal G\setminus\mathcal C^*$ on the error set indicates that suppressing the circuit mitigates model failures, while comparable scores on the clean set indicate that existing correct behavior is largely preserved. Together, these scores assess repair effectiveness and behavioral preservation.

\textbf{Circuit-Restricted Tuning.}
\label{sec:circuit_tuning}
Given the final error circuit $\mathcal{C}^{*}$, we restrict model updates to its corresponding parameters.
For each unit $u=(l,p,j)\in\mathcal C^*$, where $p\in\{\mathrm{gate},\mathrm{up},\mathrm{down}\}$, we tune the $j$-th output row of projection $p$ in layer $l$, together with its bias if present.
For each selected row, we introduce a zero-initialized trainable weight delta, together with a bias delta if present, while freezing all original model parameters.
We optimize the selected deltas on the error-correction set using the corrected responses to supervise both intermediate reasoning and final-answer prediction:
\begin{equation}
\min_{\Delta\theta(\mathcal C^*)}
-
\mathbb E_{(x,y^+)\sim\mathcal D_{\mathrm{err}}}
\left[
\frac{
\sum_t w_t
\log
p_{\theta+\Delta\theta(\mathcal C^*)}
\left(
y_t^+ \mid x,y_{<t}^+
\right)
}{
\sum_t w_t
}
\right].
\end{equation}
where $w_t$ assigns greater weight to final-answer tokens.
After training, we merge the best-performing deltas into the selected projection parameters, leaving all other parameters unchanged.

\section{Experiments}
\label{sec:experiments}


\subsection{Experimental Settings}
\label{sec:experimental-settings}

\textbf{Models.}
We evaluate RESCUE on two instruction-tuned language models with different architectures: Qwen3-8B~\citep{yang2025qwen3} and Llama-3.1-8B-Instruct~\citep{grattafiori2024llama3}. 

\textbf{Datasets.}
Our main experiments use GSM8K~\citep{cobbe2021training}.
We construct five \emph{pattern-specific} repair datasets, each with a $4:1$ train--test split.
Examples within each dataset share an underlying reasoning structure while varying in surface entities and numerical values.
We additionally construct a \emph{heterogeneous} GSM8K repair dataset with a $3:1$ train--test split.
For robustness evaluation, we create rephrased versions of five pattern-specific datasets, retaining the same split ratio.
We further evaluate cross-domain repair on MedMCQA~\citep{pal2022medmcqa} with the same setting.

\textbf{Baselines and Controls.}
We compare RESCUE with the base model, direct LoRA fine-tuning~\citep{hu2022lora}, and a \emph{supervised-only} variant that uses the circuit obtained before RL refinement and pruning. All localized circuits undergo the same ablation and circuit-restricted tuning procedures. We additionally evaluate ten density-matched random circuits as controls.

\textbf{Evaluation.}
We evaluate each localized circuit through ablation and circuit-restricted tuning.
For mathematical reasoning, we evaluate target-task performance on GSM8K~\citep{cobbe2021training} and generalization on MATH-500~\citep{hendrycks2021math,lightman2023verify}.
We assess capability preservation on ARC-Challenge~\citep{clark2018arc}, HellaSwag~\citep{zellers2019hellaswag}, WinoGrande~\citep{sakaguchi2020winogrande}, PIQA~\citep{bisk2020piqa}, TruthfulQA-MC1~\citep{lin2022truthfulqa}, the High School Mathematics and High School Statistics subsets of MMLU~\citep{hendrycks2021mmlu}, and the Multi-Step Arithmetic task from BBH~\citep{suzgun2023challenging}.
For cross-domain repair, we use MedMCQA~\citep{pal2022medmcqa} as the target and evaluate capability retention on the same eight benchmarks.
All public benchmark evaluations are conducted with the Language Model Evaluation Harness~\citep{gao2024language}, using identical configurations across models to enable comparisons.

\begin{figure}[t]
    \centering
    \includegraphics[width=\linewidth]{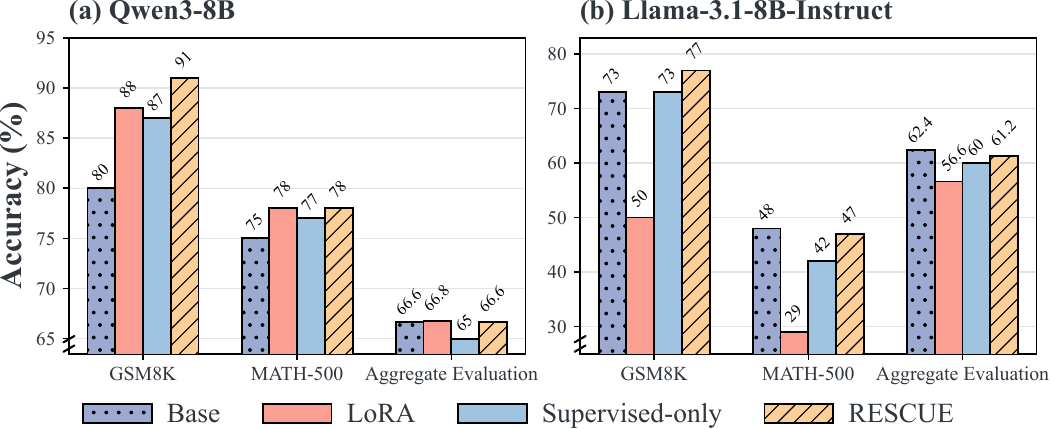}
    \caption{
Post-tuning accuracy after heterogeneous GSM8K repair; Aggregate Evaluation is the macro-average over the eight non-target benchmarks in Section~\ref{sec:experimental-settings}. 
The panels use different truncated y-axis ranges for readability, with exact percentages shown above each bar.
}
\label{fig:heterogeneous_repair}
\end{figure}

\textbf{Metrics.}
After circuit ablation, we report full-set repair accuracy on the constructed datasets and GSM8K accuracy.
After circuit-restricted tuning, we report benchmark accuracy and its absolute change from the base model.
Circuit density is measured as the percentage of selected MLP projection rows among all eligible rows.
Random-circuit ablation results are aggregated over ten density-matched samples.
Further evaluation details are provided in Appendix~\ref{app:experimental-details}.

\subsection{Heterogeneous Repair}
\label{sec:heterogeneous-repair}

We evaluate whether RESCUE can repair a heterogeneous collection of GSM8K reasoning failures rather than one error pattern. 
On the mixed repair set, we compare RESCUE with the base model, direct LoRA, and the supervised baseline. 
Figure~\ref{fig:heterogeneous_repair} reports post-tuning accuracy on GSM8K, MATH-500, and Aggregate Evaluation, defined as the macro-average over the eight non-target benchmarks.

\textbf{Target-task improvement.}
Before tuning, circuit ablation raises full-set repair accuracy from 0\% to 50.5\% on Qwen3-8B and from 6.0\% to 69.0\% on Llama-3.1-8B-Instruct, using only 1.49\% and 1.40\% of the eligible MLP rows, respectively.
Restricting tuning to their parameters then yields the highest GSM8K accuracy on both models.
On Qwen3-8B, it improves accuracy from $80\%$ to $91\%$, outperforming LoRA ($88\%$) and supervised-only ($87\%$) by 3 and 4 percentage points, respectively.
On Llama-3.1-8B-Instruct, RESCUE improves accuracy from $73\%$ to $77\%$, whereas supervised-only remains at $73\%$ and LoRA decreases to $50\%$.
The advantage over supervised-only indicates that the RESCUE pipeline---output-guided mask refinement followed by pruning---produces circuits that are more effective for downstream repair.
Circuit-restricted tuning also raises full-set repair accuracy to 53.5\% and 75.5\% on the two models, respectively.

\textbf{Generalization and capability preservation.}
RESCUE largely preserves mathematical generalization and non-target capabilities.
On Qwen3-8B, it improves MATH-500 from $75\%$ to $78\%$ and retains Aggregate Evaluation at $66.6\%$.
On Llama-3.1-8B-Instruct, it retains $47\%$ on MATH-500 and $61.2\%$ on Aggregate Evaluation, versus base scores of $48\%$ and $62.4\%$, while supervised-only falls to $42\%$ and $60.0\%$ and LoRA to $29\%$ and $56.6\%$.
Detailed results are provided in Appendix~\ref{app:heterogeneous-detailed-results}.
Overall, RESCUE has a trade-off between repairing failures and preserving capabilities.

\subsection{Pattern-Specific Error Circuits and Attribution Specificity}
\label{sec:causal_validation}

\textbf{Pattern-specific localization and repair.}
To test whether RESCUE localizes components that causally contribute to recurring failures, we independently identify and ablate circuits for five pattern-specific GSM8K repair sets per model. Table~\ref{tab:pattern_ablation} reports full-set accuracy before and after ablation, together with the proportion of selected MLP rows. Across the five independently tuned circuits, Aggregate Evaluation differs from the base model by at most 0.75 percentage points for Qwen3-8B and 1.25 points for Llama-3.1-8B-Instruct, indicating that pattern-specific repair largely preserves broader capabilities. Detailed results are provided in Appendix~\ref{app:pattern-specific-post-tuning}.


\providecolor{TableHeader}{HTML}{DCE5F1}
\providecolor{TableHighlight}{HTML}{EDF3F8}
\providecolor{TableAccent}{HTML}{315E85}

\begin{table}[t]
    \centering
    \caption{Pattern-specific circuit ablation. Circuit denotes the proportion of selected MLP rows.}
    \label{tab:pattern_ablation}
    \small
    \setlength{\tabcolsep}{4pt}
    \renewcommand{\arraystretch}{1.02}
    \resizebox{\linewidth}{!}{%
    \begin{tabular}{llcccc}
        \toprule
        \rowcolor{TableHeader}
        \textbf{Model} & \textbf{Error Pattern}
        & \textbf{Base (Acc.)}
        & \textbf{RESCUE Abl. (Acc.)}
        & \textbf{$\Delta$ (pp)}
        & \textbf{Circuit} \\
        \midrule
        \multirow{5}{*}{\textbf{Qwen3-8B}}
          & Misleading Premise Back-solving       & 2\%  & \cellcolor{TableHighlight}\textbf{93\%} & \textcolor{TableAccent}{\textbf{+91}} & 0.32\% \\
          & Dynamic State Tracking                 & 1\%  & \cellcolor{TableHighlight}\textbf{91\%} & \textcolor{TableAccent}{\textbf{+90}} & 0.31\% \\
          & Dropped-lowest Average                 & 4\%  & \cellcolor{TableHighlight}\textbf{94\%} & \textcolor{TableAccent}{\textbf{+90}} & 0.26\% \\
          & Declining Group Accumulation           & 2\%  & \cellcolor{TableHighlight}\textbf{76\%} & \textcolor{TableAccent}{\textbf{+74}} & 0.44\% \\
          & Multiplicative Increment Back-solving & 0\%  & \cellcolor{TableHighlight}\textbf{91\%} & \textcolor{TableAccent}{\textbf{+91}} & 0.39\% \\
        \midrule
        \multirow{5}{*}{\shortstack[l]{\textbf{Llama-3.1-8B-}\\\textbf{Instruct}}}
          & Cumulative Target Gap     & 1\%  & \cellcolor{TableHighlight}\textbf{95\%} & \textcolor{TableAccent}{\textbf{+94}} & 0.62\% \\
          & Scale Conversion          & 36\% & \cellcolor{TableHighlight}\textbf{92\%} & \textcolor{TableAccent}{\textbf{+56}} & 0.74\% \\
          & Percentage Risk           & 6\%  & \cellcolor{TableHighlight}\textbf{95\%} & \textcolor{TableAccent}{\textbf{+89}} & 0.60\% \\
          & Unit Price and Total Cost & 31\% & \cellcolor{TableHighlight}\textbf{95\%} & \textcolor{TableAccent}{\textbf{+64}} & 0.63\% \\
          & Fees and Change           & 24\% & \cellcolor{TableHighlight}\textbf{94\%} & \textcolor{TableAccent}{\textbf{+70}} & 0.60\% \\
        \bottomrule
    \end{tabular}%
    }
\vspace{-1.5em}
\end{table}

Ablating the localized circuits consistently corrects the corresponding failures using highly sparse interventions. On Qwen3-8B, accuracy increases from $0$--$4\%$ to $76$--$94\%$, corresponding to gains of $74$--$91$ percentage points with circuits comprising only $0.26$--$0.44\%$ of the candidate rows. On Llama-3.1-8B-Instruct, accuracy increases from $1$--$36\%$ to $92$--$95\%$, with gains of $56$--$94$ points using only $0.60$--$0.74\%$ of the candidate rows. These results provide causal evidence that sparse internal components contribute to recurring reasoning failures.

\textbf{Density-matched attribution controls.}
The gains above might still arise from applying the repair procedure to any sparse subset of the same size, which would make circuit attribution unnecessary. To test this possibility, we initialize ten per-tensor density-matched random circuit masks for two patterns per model and apply the same supervised mask-optimization and RL-refinement stages. We then ablate the resulting circuits and evaluate them on the corresponding examples subsets. All scores are ablation results; no circuit-restricted tuning is performed.

Despite selecting the same number of rows and using the same mask-optimization procedure, the random-circuit controls remain close to the base model across all four settings, with mean accuracy ranging from $0.0\%$ to $13.5\%$ and a maximum individual result of $30\%$. In contrast, RESCUE achieves $90$--$95\%$ on the same subsets. For Llama-3.1-8B-Instruct on Pattern 1, all ten random-circuit controls obtain $0\%$, whereas RESCUE reaches $95\%$. These results show that neither circuit density nor the repair procedure alone is sufficient; effective repair depends on first attributing the failure to functionally relevant model components rather than selecting arbitrary sparse subsets.


\providecolor{TableHeader}{HTML}{DCE5F1}
\providecolor{TableHighlight}{HTML}{EDF3F8}
\providecolor{TableAccent}{HTML}{315E85}

\begin{table}[t]
    \centering
    \caption{Cross-rephrase transfer of pattern-specific error circuits. Accuracy is measured over 100 rephrased examples before and after ablating the circuit localized from the original formulation.}
    \label{tab:rephrase_transfer}
    \small
    \setlength{\tabcolsep}{4pt}
    \renewcommand{\arraystretch}{1.02}
    \resizebox{\linewidth}{!}{%
    \begin{tabular}{llccc}
        \toprule
        \rowcolor{TableHeader}
        \textbf{Model} & \textbf{Error Pattern}
        & \shortstack{\textbf{Rephrased}\textbf{(Acc.)}}
        & \shortstack{\textbf{Ablation (Acc.)}}
        & \textbf{$\Delta$ (pp)} \\
        \midrule
        \multirow{5}{*}{\textbf{Qwen3-8B}}
          & Misleading Premise Back-solving       & 28\% & \cellcolor{TableHighlight}\textbf{42\%} & \textcolor{TableAccent}{\textbf{+14}} \\
          & Dynamic State Tracking                 & 17\% & \cellcolor{TableHighlight}\textbf{25\%} & \textcolor{TableAccent}{\textbf{+8}}  \\
          & Dropped-lowest Average                 & 3\%  & \cellcolor{TableHighlight}\textbf{16\%} & \textcolor{TableAccent}{\textbf{+13}} \\
          & Declining Group Accumulation           & 1\%  & \cellcolor{TableHighlight}\textbf{18\%} & \textcolor{TableAccent}{\textbf{+17}} \\
          & Multiplicative Increment Back-solving & 0\%  & \cellcolor{TableHighlight}\textbf{23\%} & \textcolor{TableAccent}{\textbf{+23}} \\
        \midrule
        \multirow{5}{*}{\shortstack[l]{\textbf{Llama-3.1-8B-}\\\textbf{Instruct}}}
          & Cumulative Target Gap     & 25\% & \cellcolor{TableHighlight}\textbf{60\%} & \textcolor{TableAccent}{\textbf{+35}} \\
          & Scale Conversion          & 35\% & \cellcolor{TableHighlight}\textbf{53\%} & \textcolor{TableAccent}{\textbf{+18}} \\
          & Percentage Risk           & 26\% & \cellcolor{TableHighlight}\textbf{92\%} & \textcolor{TableAccent}{\textbf{+66}} \\
          & Unit Price and Total Cost & 42\% & \cellcolor{TableHighlight}\textbf{94\%} & \textcolor{TableAccent}{\textbf{+52}} \\
          & Fees and Change           & 15\% & \cellcolor{TableHighlight}\textbf{75\%} & \textcolor{TableAccent}{\textbf{+60}} \\
        \bottomrule
    \end{tabular}%
    }
\vspace{-1.5em}
\end{table}

\subsection{Cross-Rephrase Transfer of Error Circuits}
\label{sec:rephrase_transfer}
\textbf{Cross-rephrase evaluation.} Pattern-specific datasets share similar surface forms, raising the concern that the localized circuits may capture template-specific lexical patterns rather than the underlying failure-producing computation. We therefore rephrase each dataset while preserving its reasoning structure and directly ablate, on the rephrased inputs, the circuit localized from the original formulation. No circuit re-localization, mask refinement, or parameter tuning is performed. Table~\ref{tab:rephrase_transfer} reports accuracy over all 100 rephrased examples before and after this same-circuit ablation.

\textbf{Results.} Same-circuit ablation improves accuracy for all ten model--pattern combinations. The gains range from 8 to 23 percentage points for Qwen3-8B, with an average of 15.0 points, and from 18 to 66 points for Llama-3.1-8B-Instruct, with an average of 46.2 points. Since the circuits are localized only from the original formulations, these consistent gains indicate that their causal effects transfer across surface reformulations. This suggests RESCUE captures relevant components linked with recurring reasoning failures rather than template-specific lexical mappings, although varying gains suggest that the same circuit need not account equally for every linguistic realization. Complementary results obtained by re-localizing circuits on the rephrased datasets appear in Appendix~\ref{app:rephrase-relocalization}.

\begin{figure}[h]
    \centering
    \includegraphics[width=\linewidth]{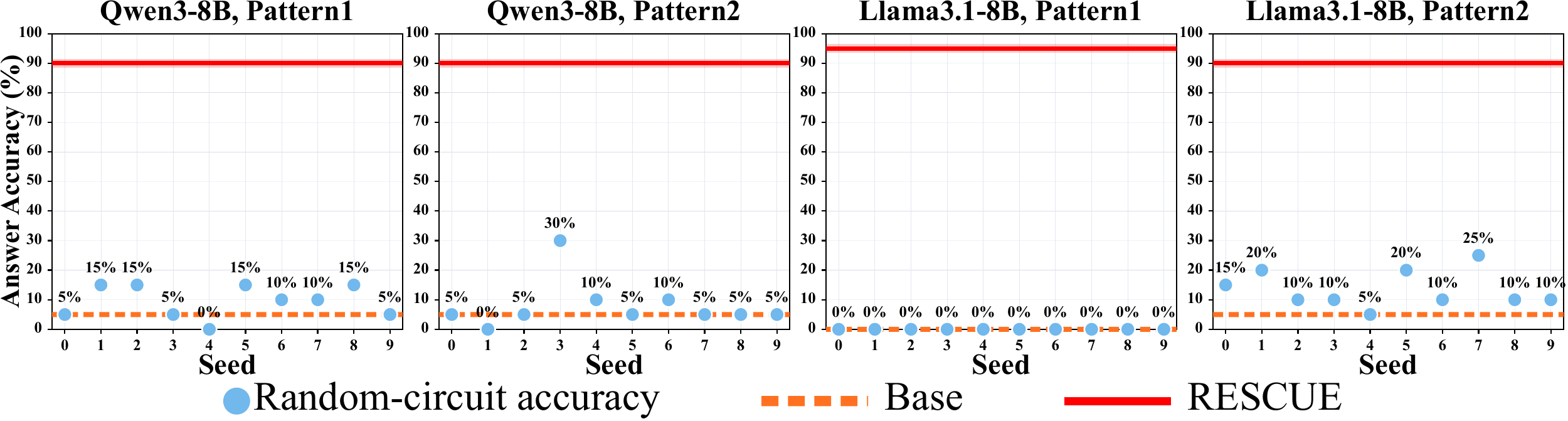}
    \caption{
    Density-matched random-circuit controls.  Pale-blue lines denote their means, while orange dashed and red solid lines denote Base and RESCUE, respectively. }  \label{fig:random_circuits}
    
\end{figure}

\subsection{Sequential Repair Preserves Earlier Repairs}
\label{sec:sequential-repair}

\begin{wrapfigure}[24]{r}{0.4\textwidth}
    \centering
    \includegraphics[width=\linewidth]{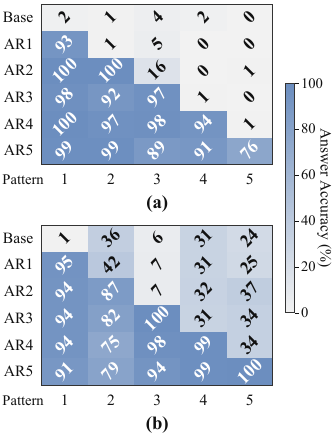}
    \caption{Full-set ablation accuracy (\%) during
    sequential repair for (a) Qwen3-8B and
    (b) Llama-3.1-8B-Instruct.
    AR1--AR5 denote repair stages; pattern order
    follows Table~1.}
    \label{fig:sequential_repair}
\end{wrapfigure}

\textbf{Sequential repair.}
In practical deployment, model failures may be discovered incrementally rather than collected in a single repair set.
We therefore apply RESCUE sequentially to five pattern-specific GSM8K datasets.
Starting from base model $M_0$, at stage $t$ we repair the next error pattern using $M_{t-1}$, yielding $M_t$.
After each stage, we evaluate $M_t$ on all five pattern sets; Figure~\ref{fig:sequential_repair} reports full-set ablation accuracy across repair stages and error patterns.

\par
\Needspace{28\baselineskip}

\textbf{Results.} Mean accuracy across the five pattern sets increases steadily from 1.8\% to 90.8\% for Qwen3-8B and from 19.6\% to 92.6\% for Llama-3.1-8B-Instruct. At their respective repair stages, the targeted patterns reach 93\%, 100\%, 97\%, 94\%, and 76\% for Qwen3-8B, and 95\%, 87\%, 100\%, 99\%, and 100\% for Llama-3.1-8B-Instruct. Earlier repairs are largely retained: after all five stages, the corresponding accuracies remain 99\%, 99\%, 89\%, 91\%, and 76\% for Qwen3-8B, and 91\%, 79\%, 94\%, 99\%, and 100\% for Llama-3.1-8B-Instruct, with a maximum later-stage decrease of only 8 percentage points for either model. These results show that RESCUE can accumulate repairs for newly discovered failures while introducing only limited interference with previous corrections. Post-tuning benchmark results after the final repair stage are provided in Appendix~\ref{app:sequential-post-tuning}.

\subsection{Cross-Domain Repair on MedMCQA}
\textbf{Cross-domain evaluation.} To evaluate cross-domain repair, we apply RESCUE and direct LoRA to heterogeneous MedMCQA repair sets for both models. Table~\ref{tab:cross-domain-repair} reports full-set repair accuracy, circuit size, and benchmark performance on MedMCQA and Aggregate Evaluation.

\textbf{Results.} Across the two models, RESCUE raises repair-set accuracy from 0--3\% to 78.5--81.0\%, outperforming LoRA (65.0--65.5\%) while selecting only 1.44--1.47\% of the candidate MLP rows. After tuning, RESCUE maintains or improves MedMCQA accuracy and better preserves aggregate performance than LoRA. These results show that RESCUE extends beyond mathematical reasoning, enabling sparse targeted repair while better preserving broader capabilities than direct fine-tuning. Full per-benchmark results are provided in Appendix~\ref{app:medmcqa-detailed-results}.


\providecolor{TableHeader}{HTML}{DCE5F1}
\providecolor{TableHighlight}{HTML}{EDF3F8}
\providecolor{TableAccent}{HTML}{315E85}

\begin{table}[t]
\centering
\caption{Cross-domain repair results on MedMCQA. Repair-set accuracy is measured after RESCUE and after LoRA. Aggregate Eval. is the mean accuracy over the eight benchmarks.}
\label{tab:cross-domain-repair}
\small
\renewcommand{\arraystretch}{1.05}
\setlength{\tabcolsep}{4pt}
\resizebox{\linewidth}{!}{%
\begin{tabular}{@{}>{\raggedright\arraybackslash}p{4.15cm}
                    >{\raggedright\arraybackslash}p{1.75cm}
                    >{\centering\arraybackslash}p{2.90cm}
                    >{\centering\arraybackslash}p{1.65cm}
                    >{\centering\arraybackslash}p{2.05cm}
                    >{\centering\arraybackslash}p{2.55cm}@{}}
\toprule
\rowcolor{TableHeader}
\multicolumn{2}{@{}c}{}
& \multicolumn{2}{c}{\textbf{Targeted Repair}}
& \multicolumn{2}{c@{}}{\textbf{Post-tuning Evaluation}} \\
\rowcolor{TableHeader}
\textbf{Model} & \textbf{Method} & \textbf{Repair-set (Acc.)}
& \textbf{Circuit} & \textbf{MedMCQA} & \textbf{Aggregate Eval.} \\
\midrule
\multirow{3}{*}{\textbf{Qwen3-8B}}
  & Base   & 3.0\%           & --      & 59\%          & 66.63\%          \\
  & LoRA   & 65.0\%          & --      & \textbf{60\%} & 64.75\%          \\
  & \cellcolor{TableHighlight}\textcolor{TableAccent}{\textbf{RESCUE}}
  & \cellcolor{TableHighlight}\textcolor{TableAccent}{\textbf{78.5\%}}
  & \cellcolor{TableHighlight}\textcolor{TableAccent}{\textbf{1.47\%}}
  & \cellcolor{TableHighlight}\textbf{60\%}
  & \cellcolor{TableHighlight}\textbf{66.75\%} \\
\addlinespace[3pt]
\midrule
\multirow{3}{*}{\textbf{Llama-3.1-8B-Instruct}}
  & Base   & 0.0\%           & --      & \textbf{59\%} & \textbf{62.38\%} \\
  & LoRA   & 65.5\%          & --      & 57\%          & 58.88\%          \\
  & \cellcolor{TableHighlight}\textcolor{TableAccent}{\textbf{RESCUE}}
  & \cellcolor{TableHighlight}\textcolor{TableAccent}{\textbf{81.0\%}}
  & \cellcolor{TableHighlight}\textcolor{TableAccent}{\textbf{1.44\%}}
  & \cellcolor{TableHighlight}\textbf{59\%}
  & \cellcolor{TableHighlight}59.75\% \\
\bottomrule
\end{tabular}
}
\vspace{-1.5em}
\end{table}

\section{Conclusion}
In this work, we introduce \textbf{RESCUE}, a framework for localizing sparse circuits associated with model errors and restricting repair to their parameters. RESCUE combines reference-based mask optimization with outcome-guided refinement over multiple model generations, followed by circuit pruning and targeted tuning. Experiments on Qwen3-8B and Llama-3.1-8B-Instruct across mathematical and medical tasks show that recurring errors are associated with highly sparse MLP circuits whose ablation improves task performance. RESCUE further outperforms the supervised-only baseline and LoRA while largely preserving non-target capabilities. Specifically, on heterogeneous GSM8K repair, RESCUE improves accuracy by 4 percentage points over supervised-only localization on both models and by up to 27 points over LoRA. Our findings extend circuit-based mechanistic interpretability beyond behavior analysis and safety interventions to targeted repair.

\section*{AI Use Statement}

Generative AI tools were used to produce the \texttt{corrected\_reasoning} field in our dataset, assist with literature retrieval, review manuscript formatting, and suggest caption and editorial revisions. All AI-assisted outputs and suggestions were reviewed by the authors. The authors take full responsibility for the final content of this work, including all text, data, results, and claims.

\section*{Reproducibility Statement}

To support reproducibility, we release our implementation in an anonymous repository. The repository includes code for supervised mask optimization, reinforcement-learning-based mask refinement, circuit pruning, circuit ablation, circuit-restricted tuning, and benchmark evaluation. Detailed data-construction procedures, prompts, experimental protocols, and evaluation metrics are provided in the Appendix. Together, these resources support reproduction of the main experiments and analyses reported in this paper.

\bibliography{iclr2027_conference}
\bibliographystyle{iclr2027_conference}

\clearpage
\appendix
\definecolor{PromptGeneration}{HTML}{5D7893}
\definecolor{PromptRephrase}{HTML}{4F786F}
\definecolor{PromptCorrection}{HTML}{71658A}
\definecolor{PromptGSM}{HTML}{465F78}
\definecolor{PromptMed}{HTML}{536F60}

\section{Experimental Details}
\label{app:experimental-details}
\subsection{Dataset Construction}
\label{app:dataset-construction}

Our experiments use paired error-correction and clean sets with complementary roles. For each model, we first run the unmodified model on candidate questions under the task-specific prompt in Appendix~\ref{app:data-construction-prompts}. Incorrect generations provide the source examples for repair, whereas correctly answered questions provide clean examples. For each incorrect generation, we retain the question, ground-truth answer, original model response, parsed prediction, and correctness label. Gemini 3.1 Pro then produces a minimally corrected reasoning trace that preserves valid intermediate steps and changes only the content needed to recover the ground-truth answer. The resulting triples of question, erroneous response, and corrected response form the error-correction set used for circuit localization and tuning. All localization and tuning data originate from the official training splits and are disjoint from benchmark evaluation data.

\paragraph{Pattern-specific GSM8K sets.}
For each model, we construct five datasets, each centered on one recurring reasoning failure. The Qwen3-8B patterns are \emph{Misleading Premise Back-solving}, \emph{Dynamic State Tracking}, \emph{Dropped-lowest Average}, \emph{Declining Group Accumulation}, and \emph{Multiplicative Increment Back-solving}. The Llama-3.1-8B-Instruct patterns are \emph{Cumulative Target Gap}, \emph{Scale Conversion}, \emph{Percentage Risk}, \emph{Unit Price and Total Cost}, and \emph{Fees and Change}.

Starting from a representative failure, Gemini 3.1 Pro generates variants that preserve the underlying sequence of mathematical operations while varying names, locations, numerical values, background scenarios, and wording. Gemini is also used to verify the generated answer and reasoning trace and to identify duplicate, invalid, or structurally inconsistent items. Samples that do not preserve the intended solution logic are excluded. Each finalized dataset contains 100 examples.

\paragraph{Rephrased sets.}
For cross-rephrase transfer, Gemini 3.1 Pro rewrites the pattern-specific questions while preserving their entities, numerical conditions, mathematical relationships, solution method, and final answer. The rewrite changes sentence structure, voice, modifier placement, and narrative organization. Each rephrased dataset contains 100 examples corresponding one-to-one with its original-formulation dataset. The main-text transfer experiment applies the circuit localized on the original formulation directly to these rephrased inputs, without re-localization or parameter tuning. Appendix~\ref{app:rephrase-relocalization} additionally reruns the complete localization procedure on two rephrased patterns per model.

\paragraph{Sequential and cross-domain sets.}
Sequential repair uses the same five pattern-specific datasets in the within-model order reported in the main paper. At each stage, the next pattern is repaired from the model state produced by the preceding stage, after which all five pattern sets are reevaluated. For cross-domain evaluation, we construct a MedMCQA repair set separately for each model. MedMCQA examples retain the multiple-choice options and ground-truth option, while their corrected references explain the selected medical answer under the same minimum-intervention principle.

\subsection{Prompt Templates}
\label{app:data-construction-prompts}

We use Gemini 3.1 Pro for three data-construction operations: generating variants of a recurring problem type, rephrasing existing questions without changing their semantics, and minimally correcting erroneous model reasoning. The following boxes provide faithful English translations of the original Chinese prompts while preserving their instructions and constraints. Answer verification and deduplication are conducted as follow-up checks within the same interaction rather than with separately fixed templates. Placeholders such as \texttt{[EXAMPLE]}, \texttt{[PROBLEMS]}, and \texttt{[SAMPLES]} are replaced at runtime with the corresponding source items.
\begin{figure}[htbp]
\centering
\begin{eninputbox}[colback=PromptGeneration!7,
colframe=PromptGeneration,
colbacktitle=PromptGeneration,
title={Pattern-Specific Problem Generation Prompt}]
\fontsize{9}{11}\selectfont
\begin{verbatim}
Given the following example problem, generate 100 similar problems. Change incidental details 
such as names, locations, and numbers, and introduce some diversity, but do not change the 
overall solution logic.
While preserving the original solution procedure, diversify the background, names, scenarios, 
numerical values, answers, and wording.
Return the results in JSON format and assign each item an ID from 1 to 100.

Example:
[EXAMPLE]
\end{verbatim}
\end{eninputbox}
\caption{English translation of the prompt for generating pattern-specific GSM8K problems.}
\label{fig:prompt_pattern_generation}
\end{figure}

\begin{figure}[t]
\centering
\begin{eninputbox}[colback=PromptRephrase!7,
colframe=PromptRephrase,
colbacktitle=PromptRephrase,
title={Problem Rephrasing Prompt}]
\fontsize{9}{11}\selectfont
\begin{verbatim}
I will provide a set of problem samples.
Rewrite only their wording and presentation without changing their meaning, logic, or solution 
method.
Thoroughly change the sentence structure, voice, modifier placement, and narrative organization 
so that each item reads like a newly written problem while exactly preserving
the original mathematical logic.
Do not freely alter any problem. Preserve the names, locations, numerical values, mathematical 
relationships,  solution logic, and answer. The rewritten problem must be fully equivalent to 
the original, rather than merely yielding the same final answer.

Problems:
[PROBLEMS]
\end{verbatim}
\end{eninputbox}
\caption{English translation of the prompt for rephrasing pattern-specific problems.}
\label{fig:prompt_rephrasing}
\end{figure}

\begin{figure*}[t]
\centering
\begin{eninputbox}[colback=PromptCorrection!7,
colframe=PromptCorrection,
colbacktitle=PromptCorrection,
title={\fontsize{10}{10}\selectfont Minimal-Intervention Correction Prompt}]
\fontsize{8.5}{10.5}\selectfont
\begin{Verbatim}[breaklines=true,breakanywhere=true]
I will next provide samples containing erroneous model outputs. Correct their erroneous 
outputs according to the minimum-intervention principle. 
Preserve every valid reasoning step and modify only the content necessary to correct the error.
Place the minimally modified response in the "corrected_reasoning" field. Return only the 
following fields in JSON format:

"id"
"question"
"ground_truth"
"generation"
"pred_answer_masked"
"is_correct_masked"
"corrected_reasoning"

Samples:
[SAMPLES]
\end{Verbatim}
\end{eninputbox}
\caption{English translation of the prompt for minimally correcting erroneous reasoning.}
\label{fig:prompt_minimal_correction}
\end{figure*}

\paragraph{Illustrative example.}
Figure~\ref{fig:minimal_correction_example} illustrates the minimum-intervention correction process. Only the erroneous calculation and its downstream result are revised, while the remaining reasoning is preserved.

\begin{figure}[t]
\centering
\begin{tcolorbox}[
    enhanced,
    width=\linewidth,
    colback=PromptCorrection!5,
    colframe=PromptCorrection,
    colbacktitle=PromptCorrection,
    boxrule=0.8pt,
    arc=2pt,
    left=5pt,
    right=5pt,
    top=4pt,
    bottom=4pt,
    title={\fontsize{10}{10}\selectfont Illustrative Minimal-Intervention Correction},
    coltitle=white,
    fonttitle=\bfseries
]
\small

\textbf{Question.}
Parker buys a \$100 string trimmer and a leaf blower costing twice as much. Both items receive a 20\% discount, after which she buys \$50 safety ear muffs. How much does she spend in total?

\medskip
\textbf{Original erroneous output.}

\begin{quote}
\small
\ldots Since Parker gets a 20\% discount, the leaf blower costs
\textcolor{red!75!black}{\$200-\$20=\$180}, while the string trimmer costs
\$100-\$20=\$80. Therefore, Parker spends
\textcolor{red!75!black}{\$180+\$80+\$50=\$310}.
\end{quote}

\textbf{Minimally corrected output.}

\begin{quote}
\small
\ldots Since Parker gets a 20\% discount, the leaf blower costs
\textcolor{blue!70!black}{\$200-\$40=\$160}, while the string trimmer costs
\$100-\$20=\$80. Therefore, Parker spends
\textcolor{blue!70!black}{\$160+\$80+\$50=\$290}.
\end{quote}

\textbf{Ground-truth answer:} \$290.
\end{tcolorbox}
\caption{Illustrative minimum-intervention correction. Colored text marks the erroneous and revised quantities; unchanged reasoning is omitted for brevity.}
\label{fig:minimal_correction_example}
\end{figure}

\paragraph{Task and reward prompts.}
The following boxes report the task-generation prompts and reasoning-evaluation prompts used during RL refinement. The task prompts are provided as system messages to the repaired model, whereas the evaluation prompts are provided to Qwen2.5-14B-Instruct, which assigns the reasoning reward. The same prompt is used for Qwen3-8B and Llama-3.1-8B-Instruct within each domain. The same output contracts are also retained across RESCUE and the corresponding comparison methods.

\begin{eninputbox}[colback=PromptGSM!7,
colframe=PromptGSM,
colbacktitle=PromptGSM,
title={\fontsize{10}{10}\selectfont GSM8K Generation System Prompt}]
\ttfamily\fontsize{9}{9.5}\selectfont
\smallskip
\begin{verbatim}
Solve the following grade-school math problem.
Use concise reasoning. Do not write an introduction.
Do not use Markdown headings.
Track the changing quantities carefully.
Your response must follow this exact format:

Reasoning:
1. <short calculation>
2. <short calculation>
3. <short calculation>
4. <short calculation>
Final Answer: <only one number>

Stop immediately after the Final Answer line.
\end{verbatim}
\end{eninputbox}

\vspace{5pt}

\begin{eninputbox}[colback=PromptGSM!7,
colframe=PromptGSM,
colbacktitle=PromptGSM,
title={\fontsize{10}{10}\selectfont GSM8K Reasoning-Evaluation Prompt}]
\ttfamily\fontsize{9}{9.5}\selectfont
\smallskip
\begin{verbatim}
You are a strict math-solution judge.
Evaluate the model response with one score chosen ONLY from {0, 0.25, 0.5, 0.75, 1.0}.
Return only one JSON object in this exact schema:
{"reasoning_score": 0.75}

Scoring rules:
- reasoning_score: 1.0 means fully correct reasoning and a consistent final answer; 
0 means wrong or nonsensical reasoning; intermediate values indicate
  partial correctness.
- Grade the visible reasoning and its consistency with the final answer.

Question:
[QUESTION]

Ground truth answer:
[GROUND_TRUTH]

Model response:
[MODEL_RESPONSE]
\end{verbatim}
\end{eninputbox}

\vspace{5pt}

\begin{eninputbox}[colback=PromptMed!7,
colframe=PromptMed,
colbacktitle=PromptMed,
title={\fontsize{10}{10}\selectfont MedMCQA Generation System Prompt}]
\ttfamily\fontsize{9}{9.5}\selectfont
\smallskip
\begin{verbatim}
Answer the medical multiple-choice question.
Use concise reasoning. Do not write an introduction.
Do not use Markdown headings.
End with exactly one final answer line in the requested format.
Stop immediately after the Final Answer line.
\end{verbatim}
\end{eninputbox}

\vspace{5pt}

\begin{eninputbox}[colback=PromptMed!7,
colframe=PromptMed,
colbacktitle=PromptMed,
title={\fontsize{10}{10}\selectfont MedMCQA Reasoning-Evaluation Prompt}]
\ttfamily\fontsize{9}{9.5}\selectfont
\smallskip
\begin{verbatim}
You are a strict medical multiple-choice reasoning judge.
Evaluate the model response with one score chosen ONLY from {0, 0.25, 0.5, 0.75, 1.0}.
Return only one JSON object in this exact schema:
{"reasoning_score": 0.75}

Scoring rules:
- reasoning_score: 1.0 means the reasoning supports the correct medical option and is 
consistent with the final answer.
- reasoning_score: 0 means wrong, unsafe, nonsensical, or inconsistent with the 
correct option.
- Intermediate values indicate partially correct reasoning.

Question:
[QUESTION]

Ground truth option:
[GROUND_TRUTH]

Model response:
[MODEL_RESPONSE]
\end{verbatim}
\end{eninputbox}

\paragraph{Prompt instantiation and reward parsing.}
For the GSM8K reasoning judge, \texttt{[QUESTION]}, \texttt{[GROUND\_TRUTH]}, and \texttt{[MODEL\_RESPONSE]} are replaced by the current problem, its numerical answer, and one sampled masked-model response. The MedMCQA judge receives the corresponding multiple-choice question, ground-truth option, and sampled response. Both judges are instructed to return a JSON object containing only \texttt{reasoning\_score}. The parsed score is discretized to one of $\{0,0.25,0.5,0.75,1.0\}$ and combined with the independently computed final-answer reward in the group-relative objective of Section~\ref{sec:circuit_discovery}.

GSM8K responses terminate with a single numerical answer, whereas MedMCQA responses terminate with the selected option label and option text. Final-answer correctness is computed independently of the reasoning judge by extracting the final-answer field and comparing it with the normalized ground truth. Separating answer correctness from reasoning quality allows partially correct reasoning to provide a graded learning signal when the final answer is wrong.

\subsection{Implementation Details}
\label{app:implementation-details}

\paragraph{Mask parameterization.}
We apply masks only to output rows of the MLP gate, up, and down projections and leave all attention modules unchanged. Each eligible row has one trainable mask logit, initialized to 0.2. The original language-model parameters remain frozen throughout circuit localization. Following Section~\ref{sec:circuit_discovery}, the forward pass uses binary keep masks, while the straight-through estimator propagates gradients through the underlying continuous mask values.

\paragraph{Supervised candidate localization.}
The supervised stage is run for 10 epochs. Corrected responses from the error-correction set provide the correction objective, while original responses from the clean set provide the preservation objective. We evaluate the binary mask during localization and retain the best feasible checkpoint rather than relying only on the final continuous mask state. At the end of this stage, all rows assigned a zero keep mask constitute the supervised candidate circuit. The supervised-only baseline uses this pre-RL, pre-pruning circuit and subsequently undergoes the same ablation and circuit-restricted tuning procedures as RESCUE.

\paragraph{Outcome-guided refinement.}
RL refinement starts from the supervised candidate mask and is run for 6 epochs. For every repair example, we sample four masked-model responses and score each response using final-answer correctness and the graded reasoning score described above. Rewards are normalized within the responses generated for the same question. The refinement stage also retains an auxiliary likelihood objective on the minimally corrected reference and a clean-set preservation objective. When the correction and clean-preservation gradients conflict, the implementation projects away the conflicting component before updating the mask. Candidate updates are accepted only when they satisfy the correction and clean-performance constraints, and the last feasible binary mask is restored when a proposed update violates them.

\paragraph{Pruning and circuit-restricted tuning.}
After RL refinement, closed rows are ranked by closure strength and progressively considered for reopening. A group is reopened only when both repair and clean-set performance remain within the specified tolerances of the refined mask. This procedure removes rows whose suppression is no longer necessary and yields the final circuit used in the reported ablation experiments. Circuit-restricted tuning then introduces zero-initialized trainable deltas only for the selected gate, up, and down projection rows. All original parameters and all rows outside the circuit remain frozen. Training uses the minimally corrected responses, with additional weight on final-answer tokens; the best-performing deltas are finally merged into the selected rows.

\paragraph{Baselines and random controls.}
Direct LoRA applies conventional low-rank fine-tuning to the same repair data without circuit localization. The supervised-only baseline omits both RL mask refinement and circuit pruning but uses the same downstream validation and restricted-tuning pipeline. For the density-matched control, we initialize ten random per-tensor masks with the same number of selected rows as the attributed circuit and apply the same supervised mask optimization and RL refinement. We then perform circuit ablation and evaluate each control on the corresponding subset.

\paragraph{Precision and reproducibility.}
All experiments use BF16 precision and a fixed random seed of 42, except that the density-matched analysis explicitly evaluates ten random mask seeds. For Qwen3-8B, the model-specific extended thinking mode is disabled so that both model families follow the visible reasoning format given in Appendix~\ref{app:data-construction-prompts}. Identical task prompts and benchmark configurations are used across methods within each model. Complete optimizer settings, checkpoint-selection thresholds, pruning tolerances, and baseline configurations are included in the accompanying code.

\subsection{Evaluation and Reporting Conventions}
\label{app:evaluation-conventions}

\paragraph{Public benchmarks.}
For GSM8K repair, we first evaluate the circuit-ablated model on GSM8K before any parameter update. After circuit-restricted tuning, we evaluate GSM8K, MATH-500, and the eight non-target benchmarks listed in Section~5.1 with the Language Model Evaluation Harness under identical task configurations. For MedMCQA repair, MedMCQA is the target public benchmark and GSM8K is reported separately as an additional cross-domain retention measure. \emph{Aggregate Evaluation} is the unweighted macro-average of ARC-Challenge, HellaSwag, WinoGrande, PIQA, TruthfulQA-MC1, MMLU High School Mathematics, MMLU High School Statistics, and BBH Multi-Step Arithmetic. GSM8K, MATH-500, and MedMCQA are not included.

For constructed GSM8K repair sets, numerical answers are extracted from the designated final-answer field and compared with normalized ground-truth answers. For MedMCQA, correctness is determined from the selected option. Public-benchmark scores follow the corresponding Language Model Evaluation Harness task implementation.

\paragraph{Circuit density.}
Circuit density is the number of selected MLP projection rows divided by the total number of eligible gate, up, and down projection rows. The denominator is computed separately for each model architecture. All percentages therefore describe a fraction of the candidate MLP-row search space.

\paragraph{Specialized controls.}
The cross-rephrase transfer experiment evaluates all 100 rephrased examples using the original-formulation circuit without re-localization, mask refinement, or tuning. The re-localization experiment in Appendix~\ref{app:rephrase-relocalization} instead reruns the complete RESCUE pipeline on the rephrased data before ablation and tuning. In sequential repair, AR1--AR5 denote the model states obtained after stages 1--5, and every state is evaluated on all five pattern sets to expose both accumulation and forgetting. In the random-circuit control, each point is an ablation accuracy from one density-matched seed; the mean summarizes the ten seeds, while the spread reflects sensitivity to random circuit initialization.

\section{Additional Experimental Results}
\label{app:additional-results}

This section provides the complete benchmark-level results that underlie the aggregate quantities reported in the main paper. Repair-set accuracy measures whether the constructed failures are corrected, whereas the public benchmarks measure the behavior of the post-tuning model outside those examples. Aggregate Evaluation is the unweighted mean over the eight non-target benchmarks defined in Appendix~\ref{app:evaluation-conventions}.

\paragraph{Aggregate capability retention.}
To summarize capability preservation across experiments, we compute the retention ratio between the post-tuning Aggregate Evaluation of RESCUE and that of the corresponding base model. Specifically,
\[
\mathrm{Retention}
=
\frac{1}{20}
\sum_{s=1}^{20}
\frac{A_{s}^{\mathrm{RESCUE}}}
     {A_{s}^{\mathrm{Base}}}
\times 100\%,
\]
where $A_{s}^{\mathrm{RESCUE}}$ and $A_{s}^{\mathrm{Base}}$ denote Aggregate Evaluation in repair setting $s$ for RESCUE and its corresponding base model. The 20 settings comprise two heterogeneous GSM8K repairs, ten independently tuned pattern-specific repairs, four re-localized rephrase repairs, two final models after sequential repair, and two MedMCQA repairs. Each model--setting pair is weighted equally. This calculation yields an average retention of 99.4\% relative to the corresponding base models.
\label{app:aggregate-retention}

\subsection{Detailed Results on Non-Target Benchmarks}
\label{app:heterogeneous-detailed-results}

\paragraph{Purpose and protocol.}
The heterogeneous GSM8K experiment asks whether one circuit can be localized from a mixture of failure types and then tuned without broadly degrading unrelated capabilities. Figure~2 compresses the eight non-target tasks into Aggregate Evaluation. Table~\ref{tab:heterogeneous_detailed_results} expands this average and reports the underlying benchmark values for the base model, direct LoRA, supervised-only localization, and the complete RESCUE pipeline. Supervised-only and RESCUE use their respective localized circuits for circuit-restricted tuning, while LoRA updates low-rank adapters without a localized circuit.


\providecolor{TableHeader}{HTML}{DCE5F1}
\providecolor{TableHighlight}{HTML}{EDF3F8}
\providecolor{TableAccent}{HTML}{315E85}

\begin{table*}[t]
    \centering
    \caption{Detailed post-tuning accuracy (\%) after heterogeneous GSM8K repair; Aggregate Eval. is the macro-average over the eight non-target benchmarks.}
    \label{tab:heterogeneous_detailed_results}
    \small
    \setlength{\tabcolsep}{3.2pt}
    \renewcommand{\arraystretch}{1.04}
    \resizebox{\textwidth}{!}{%
    \begin{tabular}{llcccc}
        \toprule
        \rowcolor{TableHeader}
        \textbf{Model} & \textbf{Benchmark}
        & \textbf{Base}
        & \textbf{LoRA}
        & \textbf{Supervised-only}
        & \cellcolor{TableHighlight}\textcolor{TableAccent}{\textbf{RESCUE}} \\
        \midrule
        \multirow{9}{*}{\textbf{Qwen3-8B}}
          & ARC-C           & 56    & 57    & 57    & \cellcolor{TableHighlight}57    \\
          & HellaSwag       & 75    & 76    & 75    & \cellcolor{TableHighlight}76    \\
          & WinoGrande      & 68    & 67    & 68    & \cellcolor{TableHighlight}67    \\
          & PIQA            & 78    & 77    & 78    & \cellcolor{TableHighlight}77    \\
          & TruthfulQA      & 36    & 36    & 36    & \cellcolor{TableHighlight}36    \\
          & MMLU-Math       & 50    & 53    & 55    & \cellcolor{TableHighlight}50    \\
          & MMLU-Stats      & 72    & 74    & 74    & \cellcolor{TableHighlight}73    \\
          & BBH             & 98    & 94    & 77    & \cellcolor{TableHighlight}97    \\
          & Aggregate Eval. & 66.63 & 66.75 & 65.00 & \cellcolor{TableHighlight}66.63 \\
        \midrule
        \multirow{9}{*}{\shortstack[l]{\textbf{Llama-3.1-8B-}\\\textbf{Instruct}}}
          & ARC-C           & 56    & 51    & 53    & \cellcolor{TableHighlight}56    \\
          & HellaSwag       & 80    & 78    & 79    & \cellcolor{TableHighlight}80    \\
          & WinoGrande      & 74    & 74    & 73    & \cellcolor{TableHighlight}73    \\
          & PIQA            & 81    & 80    & 81    & \cellcolor{TableHighlight}81    \\
          & TruthfulQA      & 38    & 37    & 39    & \cellcolor{TableHighlight}38    \\
          & MMLU-Math       & 42    & 43    & 39    & \cellcolor{TableHighlight}41    \\
          & MMLU-Stats      & 55    & 52    & 54    & \cellcolor{TableHighlight}53    \\
          & BBH             & 73    & 38    & 62    & \cellcolor{TableHighlight}68    \\
          & Aggregate Eval. & 62.38 & 56.63 & 60.00 & \cellcolor{TableHighlight}61.25 \\
        \bottomrule
    \end{tabular}%
    }
\end{table*}

\paragraph{Ablation and circuit compactness.}
Before circuit-restricted tuning, suppressing the heterogeneous circuit improves full-set repair accuracy from 0\% to 50.5\% on Qwen3-8B and from 6.0\% to 69\% on Llama-3.1-8B-Instruct. The corresponding circuits contain only 1.49\% and 1.40\% of the eligible MLP rows, respectively. On the public GSM8K benchmark, ablation changes accuracy from 80\% to 84\% for Qwen3-8B and from 73\% to 76\% for Llama-3.1-8B-Instruct. These pre-tuning results establish that suppressing the localized rows already mitigates the mixed failure distribution.

\paragraph{Qwen3-8B.}
RESCUE exactly retains the base Aggregate Evaluation of 66.63\%. Its individual changes relative to the base model are small and balanced: ARC-C and HellaSwag each increase by one point; WinoGrande and PIQA each decrease by one point; TruthfulQA and MMLU-Math remain unchanged; MMLU-Stats increases by one point; and BBH decreases by one point. LoRA obtains a slightly higher aggregate of 66.75\%, but its benchmark profile is less uniform, including a four-point reduction on BBH that is offset by improvements on the two MMLU subsets. The supervised-only circuit reaches 65.00\%, with its largest reduction occurring on BBH.

\paragraph{Llama-3.1-8B-Instruct.}
The capability-preservation differences are larger on Llama-3.1-8B-Instruct. RESCUE reaches 61.25\%, compared with 62.38\% for the base model, 60.00\% for supervised-only localization, and 56.63\% for LoRA. RESCUE exactly preserves ARC-C, HellaSwag, PIQA, and TruthfulQA, while changing WinoGrande and MMLU-Math by one point and MMLU-Stats by two points. Its largest reduction is five points on BBH. By comparison, LoRA decreases BBH from 73\% to 38\%, which accounts for much of its aggregate loss, while supervised-only reaches 62\% on BBH.

\paragraph{Summary.}
Across both architectures, heterogeneous repair does not affect all non-target tasks uniformly. Nevertheless, RESCUE either matches the base aggregate or remains substantially closer to it than the alternative repair pipelines. Together with the higher GSM8K accuracy in the main paper, the detailed results support the conclusion that outcome-guided mask refinement and pruning improve targeted repair without requiring broad parameter updates.

\subsection{Post-Tuning Results for Pattern-Specific Repair}
\label{app:pattern-specific-post-tuning}

\paragraph{Purpose and protocol.}
The main paper establishes a causal effect by ablating each pattern-specific circuit. Here we evaluate what happens after the same circuit is used as the trainable parameter subset. Each of the five circuits is tuned independently from the corresponding base model. We then evaluate the resulting model on GSM8K, MATH-500, and the eight non-target benchmarks. Table~\ref{tab:pattern_specific_post_tuning} reports all ten independently tuned models together with their respective base models.


\providecolor{TableHeader}{HTML}{DCE5F1}
\providecolor{TableHighlight}{HTML}{EDF3F8}
\providecolor{TableAccent}{HTML}{315E85}

\begin{table*}[t]
    \centering
    \caption{Post-tuning accuracy (\%) after independently repairing each pattern-specific GSM8K dataset; Aggregate Eval. is the macro-average over the eight non-target benchmarks.}
    \label{tab:pattern_specific_post_tuning}
    \normalsize
    \setlength{\tabcolsep}{1.8pt}
    \renewcommand{\arraystretch}{1.18}
    \resizebox{\textwidth}{!}{%
    \begin{tabular}{llcccccc}
        \toprule
        \rowcolor{TableHeader}
        \textbf{Model} & \textbf{Benchmark}
        & \textbf{Base}
        & \textbf{\shortstack{Misleading Premise\\Back-solving}}
        & \textbf{\shortstack{Dynamic State\\Tracking}}
        & \textbf{\shortstack{Dropped-lowest\\Average}}
        & \textbf{\shortstack{Declining Group\\Accumulation}}
        & \textbf{\shortstack{Multiplicative Increment\\Back-solving}} \\
        \midrule
        \multirow{11}{*}{\textbf{Qwen3-8B}}
          & GSM8K           & 80    & 80    & 81    & 80    & 86    & 82    \\
          & MATH-500        & 75    & 76    & 75    & 77    & 74    & 75    \\
          & ARC-C           & 56    & 56    & 56    & 57    & 55    & 58    \\
          & HellaSwag       & 75    & 75    & 75    & 75    & 75    & 75    \\
          & WinoGrande      & 68    & 69    & 68    & 68    & 67    & 68    \\
          & PIQA            & 78    & 77    & 78    & 77    & 77    & 77    \\
          & TruthfulQA      & 36    & 36    & 36    & 36    & 36    & 37    \\
          & MMLU-Math       & 50    & 51    & 52    & 51    & 52    & 51    \\
          & MMLU-Stats      & 72    & 72    & 77    & 73    & 70    & 74    \\
          & BBH             & 98    & 98    & 97    & 97    & 98    & 97    \\
          & Aggregate Eval. & \cellcolor{TableHighlight}66.63
                            & \cellcolor{TableHighlight}66.75
                            & \cellcolor{TableHighlight}67.38
                            & \cellcolor{TableHighlight}66.75
                            & \cellcolor{TableHighlight}66.25
                            & \cellcolor{TableHighlight}67.13 \\
        \midrule
        \rowcolor{TableHeader}
        \textbf{Model} & \textbf{Benchmark}
        & \textbf{Base}
        & \textbf{\shortstack{Cumulative Target\\Gap}}
        & \textbf{\shortstack{Scale\\Conversion}}
        & \textbf{\shortstack{Percentage\\Risk}}
        & \textbf{\shortstack{Unit Price and\\Total Cost}}
        & \textbf{\shortstack{Fees and\\Change}} \\
        \midrule
        \multirow{11}{*}{\shortstack[l]{\textbf{Llama-3.1-8B-}\\\textbf{Instruct}}}
          & GSM8K           & 73    & 73    & 75    & 75    & 74    & 73    \\
          & MATH-500        & 48    & 46    & 47    & 48    & 46    & 46    \\
          & ARC-C           & 56    & 55    & 55    & 56    & 56    & 56    \\
          & HellaSwag       & 80    & 80    & 80    & 79    & 79    & 80    \\
          & WinoGrande      & 74    & 74    & 73    & 74    & 74    & 73    \\
          & PIQA            & 81    & 81    & 81    & 82    & 81    & 81    \\
          & TruthfulQA      & 38    & 38    & 38    & 38    & 38    & 37    \\
          & MMLU-Math       & 42    & 40    & 42    & 43    & 42    & 40    \\
          & MMLU-Stats      & 55    & 55    & 55    & 54    & 53    & 52    \\
          & BBH             & 73    & 74    & 76    & 75    & 70    & 70    \\
          & Aggregate Eval. & \cellcolor{TableHighlight}62.38
                            & \cellcolor{TableHighlight}62.13
                            & \cellcolor{TableHighlight}62.50
                            & \cellcolor{TableHighlight}62.63
                            & \cellcolor{TableHighlight}61.63
                            & \cellcolor{TableHighlight}61.13 \\
        \bottomrule
    \end{tabular}%
    }
\end{table*}

\paragraph{GSM8K before and after tuning.}
The public GSM8K benchmark is evaluated both immediately after circuit ablation and after circuit-restricted tuning. In the Qwen3-8B pattern order shown in Table~1, the post-ablation scores are 80\%, 82\%, 80\%, 82\%, and 82\%, compared with a base accuracy of 80\%; the corresponding post-tuning scores are 80\%, 81\%, 80\%, 86\%, and 82\%. For Llama-3.1-8B-Instruct, the post-ablation scores are 75\%, 73\%, 74\%, 74\%, and 74\%, versus a 73\% base accuracy, and the post-tuning scores are 73\%, 75\%, 75\%, 74\%, and 73\%.

\paragraph{Qwen3-8B.}
Aggregate Evaluation ranges from 66.25\% to 67.38\%, compared with 66.63\% for the base model. The largest positive difference is obtained after tuning the Dynamic State Tracking circuit ($+0.75$ points), whereas the largest reduction is produced by Declining Group Accumulation ($-0.38$ points). GSM8K remains at or above its 80\% base value for every circuit and reaches 86\% after tuning Declining Group Accumulation. MATH-500 remains within 74--77\%, compared with 75\% for the base model.

\paragraph{Llama-3.1-8B-Instruct.}
Aggregate Evaluation ranges from 61.13\% to 62.63\%, versus 62.38\% for the base model. Cumulative Target Gap and Scale Conversion remain within 0.25 and 0.12 points of the base aggregate, respectively, while Percentage Risk is 0.25 points higher. The two lowest aggregates occur for Unit Price and Total Cost and Fees and Change, driven mainly by lower BBH and MMLU-Stats results. GSM8K remains between 73\% and 75\%, and MATH-500 remains between 46\% and 48\%. Thus, the model-specific repair effect remains localized even for circuits whose ablation changes full-set repair accuracy by more than 50 percentage points.

\paragraph{Summary.}
Across the ten independently repaired models, the non-target aggregate differs from the corresponding base by at most 0.75 points for Qwen3-8B and 1.25 points for Llama-3.1-8B-Instruct. These results complement the ablation evidence in the main paper: the localized circuits are not only causally associated with the recurring failures, but can also serve as restricted update regions that correct those failures without inducing the broad shifts expected from unconstrained fine-tuning.

\begin{figure}[t]
    \centering
\includegraphics[width=\linewidth]{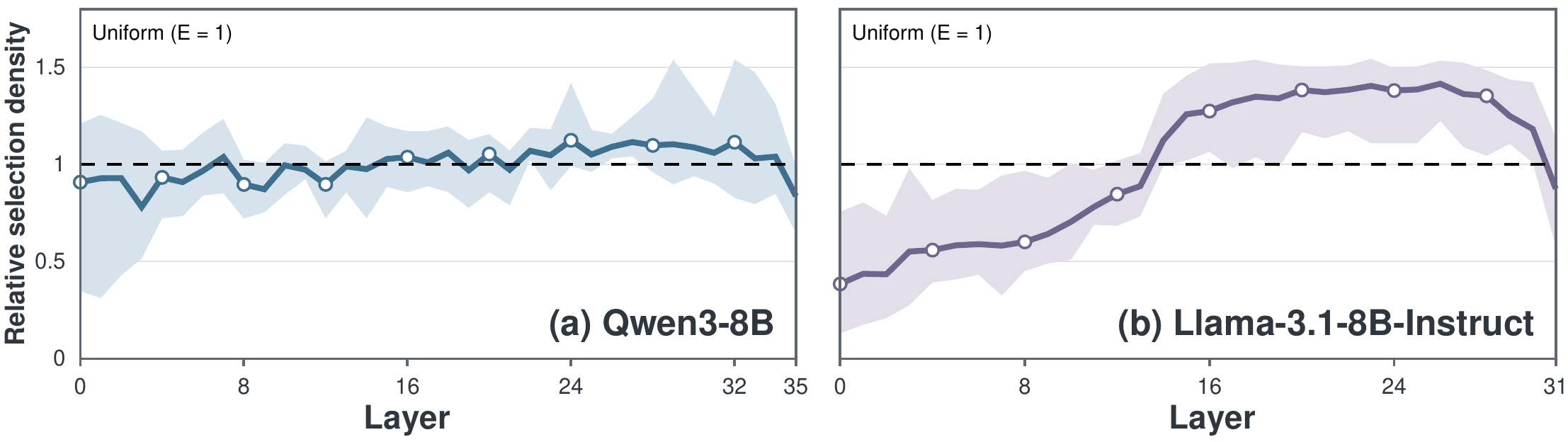}
\caption{Layer-wise enrichment of localized error-circuit rows, normalized by each circuit's overall selection density. Curves and shading show the mean and min--max range across five pattern-specific circuits; $E=1$ denotes no enrichment.}
\label{fig:layerwise-circuit-enrichment}
\end{figure}

\paragraph{Layer-wise circuit organization.}
For layer $\ell$, we define the enrichment score as
$E_\ell=(k_\ell/n_\ell)/(K/N)$, where $k_\ell$ and $n_\ell$
are the selected and eligible rows in layer $\ell$, and $K$ and
$N$ are their model-wide totals. Figure~\ref{fig:layerwise-circuit-enrichment}
shows that Qwen3-8B circuits are relatively diffuse, whereas
Llama-3.1-8B-Instruct circuits are enriched in middle-to-late layers.
Specifically, 65.7\% of the selected rows lie in the latter half of
Llama-3.1-8B-Instruct, compared with 52.6\% for Qwen3-8B, indicating
model-specific depth organization.

\subsection{Circuit Re-localization on Rephrased Inputs}
\label{app:rephrase-relocalization}

\paragraph{Purpose and protocol.}
The main-text cross-rephrase experiment applies circuits localized from the original formulations directly to all five rephrased pattern sets per model. That experiment isolates transfer: it performs no re-localization, additional mask refinement, or circuit-restricted tuning on the rewritten inputs. As a complementary and more computationally intensive analysis, we rerun the complete RESCUE pipeline on two rephrased patterns per model. The evaluated Qwen3-8B patterns are Misleading Premise Back-solving and Dynamic State Tracking; the Llama-3.1-8B-Instruct patterns are Cumulative Target Gap and Percentage Risk. Table~\ref{tab:rephrase_relocalization} compares direct circuit transfer with re-localization and reports the downstream post-tuning results.


\providecolor{TableHeader}{HTML}{DCE5F1}
\providecolor{TableHighlight}{HTML}{EDF3F8}
\providecolor{TableAccent}{HTML}{315E85}

\begin{table*}[htbt]
    \centering
    \caption{Re-localization on rephrased pattern-specific datasets. Transferred Abl. applies the original circuit, whereas Re-localized Abl. uses a circuit localized on the rephrased data; the bottom panel reports post-tuning accuracy.}
    \label{tab:rephrase_relocalization}
    \small
    \renewcommand{\arraystretch}{1.04}

    \textbf{(a) Rephrased full-set ablation}\par\vspace{2pt}
    \setlength{\tabcolsep}{4.8pt}
    \resizebox{\textwidth}{!}{%
    \begin{tabular}{llcc}
        \toprule
        \rowcolor{TableHeader}
        \textbf{Model} &
        & \textbf{\shortstack{Misleading Premise\\Back-solving}}
        & \textbf{\shortstack{Dynamic State\\Tracking}} \\
        \midrule
        \multirow{5}{*}{\textbf{Qwen3-8B}}
          & Rephrased Base          & 28\% & 17\% \\
          & Transferred Abl.        & 42\% & 25\% \\
          & Re-localized Abl.       & \cellcolor{TableHighlight}\textbf{95\%}
                                     & \cellcolor{TableHighlight}\textbf{95\%} \\
          & $\Delta$ vs. Base (pp)  & \textcolor{TableAccent}{\textbf{+67}}
                                     & \textcolor{TableAccent}{\textbf{+78}} \\
          & Circuit                 & 0.68\% & 0.69\% \\
        \midrule
        \rowcolor{TableHeader}
        \textbf{Model} &
        & \textbf{\shortstack{Cumulative Target\\Gap}}
        & \textbf{Percentage Risk} \\
        \midrule
        \multirow{5}{*}{\shortstack[l]{\textbf{Llama-3.1-8B-}\\\textbf{Instruct}}}
          & Rephrased Base          & 25\% & 26\% \\
          & Transferred Abl.        & 60\% & 92\% \\
          & Re-localized Abl.       & \cellcolor{TableHighlight}\textbf{94\%}
                                     & \cellcolor{TableHighlight}\textbf{96\%} \\
          & $\Delta$ vs. Base (pp)  & \textcolor{TableAccent}{\textbf{+69}}
                                     & \textcolor{TableAccent}{\textbf{+70}} \\
          & Circuit                 & 0.65\% & 0.65\% \\
        \bottomrule
    \end{tabular}%
    }

    \vspace{6pt}
    \textbf{(b) Post-tuning evaluation}\par\vspace{2pt}
    \setlength{\tabcolsep}{3.2pt}
    \begin{tabular}{llccc}
        \toprule
        \rowcolor{TableHeader}
        \textbf{Model} &
        & \textbf{Base}
        & \textbf{\shortstack{Misleading Premise\\Back-solving}}
        & \textbf{\shortstack{Dynamic State\\Tracking}} \\
        \midrule
        \multirow{11}{*}{\textbf{Qwen3-8B}}
          & GSM8K           & 80    & 80    & 81    \\
          & MATH-500        & 75    & 75    & 77    \\
          & ARC-C           & 56    & 56    & 57    \\
          & HellaSwag       & 75    & 75    & 74    \\
          & WinoGrande      & 68    & 69    & 70    \\
          & PIQA            & 78    & 78    & 77    \\
          & TruthfulQA      & 36    & 36    & 37    \\
          & MMLU-Math       & 50    & 51    & 50    \\
          & MMLU-Stats      & 72    & 72    & 71    \\
          & BBH             & 98    & 97    & 98    \\
          & Aggregate Eval. & \cellcolor{TableHighlight}66.63
                            & \cellcolor{TableHighlight}66.75
                            & \cellcolor{TableHighlight}66.75 \\
        \midrule
        \rowcolor{TableHeader}
        \textbf{Model} &
        & \textbf{Base}
        & \textbf{\shortstack{Cumulative Target\\Gap}}
        & \textbf{\shortstack{Percentage\\Risk}} \\
        \midrule
        \multirow{11}{*}{\shortstack[l]{\textbf{Llama-3.1-8B-}\\\textbf{Instruct}}}
          & GSM8K           & 73    & 74    & 73    \\
          & MATH-500        & 48    & 44    & 45    \\
          & ARC-C           & 56    & 55    & 56    \\
          & HellaSwag       & 80    & 80    & 79    \\
          & WinoGrande      & 74    & 73    & 74    \\
          & PIQA            & 81    & 81    & 81    \\
          & TruthfulQA      & 38    & 38    & 38    \\
          & MMLU-Math       & 42    & 41    & 41    \\
          & MMLU-Stats      & 55    & 53    & 53    \\
          & BBH             & 73    & 72    & 74    \\
          & Aggregate Eval. & \cellcolor{TableHighlight}62.38
                            & \cellcolor{TableHighlight}61.63
                            & \cellcolor{TableHighlight}62.00 \\
        \bottomrule
    \end{tabular}%
\end{table*}

\paragraph{Ablation after re-localization.}
The re-localized circuits raise full-set accuracy to 94--96\% across all four settings while selecting only 0.65--0.69\% of the candidate MLP rows. On Qwen3-8B, direct transfer raises the two rephrased sets from 28\% to 42\% and from 17\% to 25\%, whereas re-localization reaches 95\% on both. Re-localization therefore adds 53 and 70 points over direct transfer. On Llama-3.1-8B-Instruct, direct transfer is already stronger, reaching 60\% for Cumulative Target Gap and 92\% for Percentage Risk; re-localization raises these values to 94\% and 96\%, adding 34 and 4 points. The small four-point increment for Percentage Risk is not a failure of re-localization: the original circuit already transfers almost completely to that rephrased distribution.

\paragraph{Post-tuning preservation.}
For Qwen3-8B, the two re-localized circuits obtain Aggregate Evaluation scores of 66.75\%, compared with 66.63\% for the base model. GSM8K remains at 80--81\%, and MATH-500 remains at 75--77\%. For Llama-3.1-8B-Instruct, the aggregates are 61.63\% and 62.00\%, compared with 62.38\% for the base model; GSM8K remains at 73--74\%, while MATH-500 decreases from 48\% to 44--45\%. Across the eight aggregate components, the changes are small and do not exhibit a consistent collapse on any shared benchmark.

\paragraph{Interpretation and scope.}
Direct transfer and re-localization answer different questions. Direct transfer tests whether an originally localized mechanism remains causally relevant after surface reformulation, while re-localization tests whether RESCUE can rediscover a compact circuit when given the rewritten distribution itself. The results support both claims, but the re-localization analysis covers only two patterns per model and should be interpreted as a complementary robustness check rather than an exhaustive evaluation over every rephrased pattern.

\subsection{Post-Tuning Results after Sequential Repair}
\label{app:sequential-post-tuning}

\begin{figure}[t]
\centering
\includegraphics[width=0.72\linewidth]{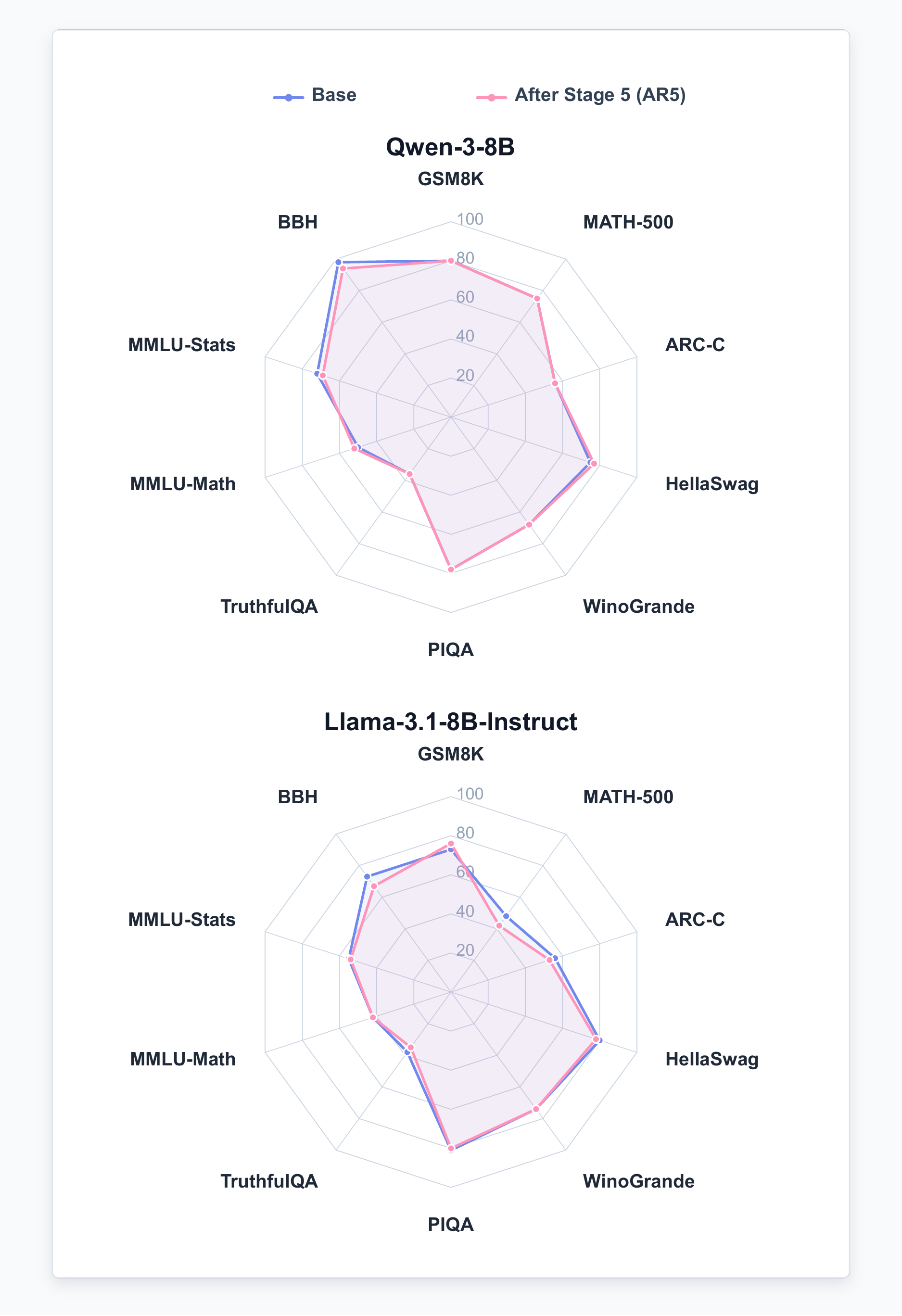}
\caption{Benchmark accuracy for the base models and their final states after five sequential repairs (AR5).}
\label{fig:sequential_post_tuning}
\end{figure}

\paragraph{Purpose and protocol.}
The main sequential-repair experiment evaluates whether errors can be repaired as they are discovered, without restarting from the original model or erasing earlier corrections. Beginning with the base model, each stage localizes and repairs the next pattern using the model state inherited from the preceding stage. The main-text heatmap evaluates all five pattern sets after every stage. Figure~\ref{fig:sequential_post_tuning} complements this within-repair-set analysis by comparing the public-benchmark performance of the base model with the final model obtained after all five stages (AR5).

\paragraph{Qwen3-8B.}
After five stages, GSM8K and MATH-500 remain unchanged at 80\% and 75\%, respectively. Aggregate Evaluation changes only from 66.63\% to 66.25\%. At the benchmark level, HellaSwag and MMLU-Math each improve by two points, ARC-C, WinoGrande, PIQA, and TruthfulQA remain unchanged, MMLU-Stats decreases by three points, and BBH decreases by four points. The small aggregate difference therefore reflects offsetting benchmark-level changes rather than a uniform downward shift.

\paragraph{Llama-3.1-8B-Instruct.}
GSM8K increases from 73\% to 76\% after sequential repair, while MATH-500 decreases from 48\% to 42\%. Aggregate Evaluation decreases by two points, from 62.38\% to 60.38\%. WinoGrande and MMLU-Math remain unchanged; PIQA and MMLU-Stats decrease by one point; HellaSwag decreases by two; ARC-C and TruthfulQA decrease by three; and BBH decreases by six. The larger change relative to Qwen3-8B shows that capability retention under repeated updates is architecture- and benchmark-dependent.

\paragraph{Interpretation.}
The main-text heatmap shows that the final AR5 models retain high accuracy on the previously repaired pattern sets, with maximum later-stage decreases of only eight points. Figure~\ref{fig:sequential_post_tuning} adds an important qualification: retaining prior repairs does not imply that every unrelated benchmark is exactly invariant. Nevertheless, five successive circuit-restricted updates preserve the Qwen3-8B aggregate almost completely and limit the Llama-3.1-8B-Instruct aggregate reduction to two points, while maintaining the accumulated repair effect. This supports sequential deployment while making the remaining model-dependent interference explicit.


\providecolor{TableHeader}{HTML}{DCE5F1}
\providecolor{TableHighlight}{HTML}{EDF3F8}
\providecolor{TableAccent}{HTML}{315E85}

\begin{table*}[t]
    \centering
    \caption{Detailed post-tuning accuracy (\%) for MedMCQA repair. Aggregate Eval. averages the eight benchmarks from ARC-C through BBH and excludes GSM8K.}
    \label{tab:medmcqa_detailed_results}
    \small
    \renewcommand{\arraystretch}{1.04}
    \setlength{\tabcolsep}{4.0pt}
    \resizebox{\textwidth}{!}{%
    \begin{tabular}{llccc}
        \toprule
        \rowcolor{TableHeader}
        \textbf{Model} & \textbf{Benchmark}
        & \textbf{Base}
        & \textbf{LoRA}
        & \cellcolor{TableHighlight}\textcolor{TableAccent}{\textbf{RESCUE}} \\
        \midrule
        \multirow{11}{*}{\textbf{Qwen3-8B}}
          & MedMCQA         & 59          & \textbf{60} & \cellcolor{TableHighlight}\textbf{60} \\
          & GSM8K           & 80          & 85          & \cellcolor{TableHighlight}\textbf{86} \\
          & ARC-C           & 56          & 57          & \cellcolor{TableHighlight}\textbf{58} \\
          & HellaSwag       & 75          & \textbf{76} & \cellcolor{TableHighlight}\textbf{76} \\
          & WinoGrande      & 68          & 68          & \cellcolor{TableHighlight}\textbf{69} \\
          & PIQA            & 78          & 76          & \cellcolor{TableHighlight}\textbf{79} \\
          & TruthfulQA      & 36          & 36          & \cellcolor{TableHighlight}\textbf{37} \\
          & MMLU-Math       & \textbf{50} & 49          & \cellcolor{TableHighlight}\textbf{50} \\
          & MMLU-Stats      & \textbf{72} & 69          & \cellcolor{TableHighlight}71 \\
          & BBH             & \textbf{98} & 87          & \cellcolor{TableHighlight}94 \\
          & Aggregate Eval. & 66.63       & 64.75       & \cellcolor{TableHighlight}\textcolor{TableAccent}{\textbf{66.75}} \\
        \midrule
        \multirow{11}{*}{\shortstack[l]{\textbf{Llama-3.1-8B-}\\\textbf{Instruct}}}
          & MedMCQA         & \textbf{59}    & 57 & \cellcolor{TableHighlight}\textbf{59} \\
          & GSM8K           & \textbf{73}    & 71 & \cellcolor{TableHighlight}72 \\
          & ARC-C           & \textbf{56}    & 53 & \cellcolor{TableHighlight}52 \\
          & HellaSwag       & \textbf{80}    & 78 & \cellcolor{TableHighlight}78 \\
          & WinoGrande      & \textbf{74}    & 72 & \cellcolor{TableHighlight}72 \\
          & PIQA            & \textbf{81}    & 80 & \cellcolor{TableHighlight}80 \\
          & TruthfulQA      & \textbf{38}    & 36 & \cellcolor{TableHighlight}35 \\
          & MMLU-Math       & \textbf{42}    & 36 & \cellcolor{TableHighlight}40 \\
          & MMLU-Stats      & \textbf{55}    & 48 & \cellcolor{TableHighlight}52 \\
          & BBH             & \textbf{73}    & 68 & \cellcolor{TableHighlight}69 \\
          & Aggregate Eval. & \textbf{62.38} & 58.88 & \cellcolor{TableHighlight}\textcolor{TableAccent}{59.75} \\
        \bottomrule
    \end{tabular}%
    }
\end{table*}

\subsection{Detailed Results for MedMCQA Repair}
\label{app:medmcqa-detailed-results}

\paragraph{Purpose and protocol.}
The MedMCQA experiment tests whether the localization-and-repair procedure extends beyond grade-school mathematical reasoning. The main paper reports repair-set accuracy, the public MedMCQA benchmark, circuit size, and Aggregate Evaluation for the base model, LoRA, and RESCUE. Table~\ref{tab:medmcqa_detailed_results} expands the aggregate into its eight constituent benchmarks and separately reports GSM8K as an additional cross-domain retention measure. The supervised-only baseline is not included in this setting because it was not run for the completed MedMCQA experiments.

\paragraph{Qwen3-8B.}
RESCUE and LoRA both obtain 60\% on the MedMCQA benchmark, compared with 59\% for the base model. Their non-target behavior differs substantially, however. RESCUE obtains an Aggregate Evaluation of 66.75\%, slightly above the base model's 66.63\%, while LoRA decreases to 64.75\%. Relative to the base model, RESCUE improves ARC-C by two points and HellaSwag, WinoGrande, PIQA, and TruthfulQA by one point, preserves MMLU-Math, decreases MMLU-Stats by one point, and decreases BBH by four points. LoRA's lower aggregate is driven mainly by reductions of three points on MMLU-Stats and eleven points on BBH. GSM8K increases from 80\% to 86\% under RESCUE and to 85\% under LoRA.

\paragraph{Llama-3.1-8B-Instruct.}
RESCUE preserves the base MedMCQA accuracy of 59\%, whereas LoRA reaches 57\%. Aggregate Evaluation decreases from 62.38\% to 59.75\% for RESCUE and to 58.88\% for LoRA. Thus, RESCUE does not fully preserve the base aggregate on this model, but remains 0.87 points above direct LoRA while achieving two points higher MedMCQA accuracy. The RESCUE changes relative to the base model range from one point on PIQA to four points on ARC-C and BBH. GSM8K changes only from 73\% to 72\%, compared with 71\% under LoRA.

\end{document}